\documentclass{article}

\usepackage{amsmath,amsfonts,bm}

\def\eqreff#1{(\ref{#1})}
\def\eqref#1{equation~(\ref{#1})}

\def\1{\bm{1}}

\DeclareMathAlphabet{\mathsfit}{\encodingdefault}{\sfdefault}{m}{sl}
\SetMathAlphabet{\mathsfit}{bold}{\encodingdefault}{\sfdefault}{bx}{n}

\DeclareMathOperator*{\arginf}{arg\,inf}

\usepackage{microtype}
\usepackage{graphicx}
\usepackage{subfigure}
\usepackage{booktabs} 

\usepackage{hyperref}

\newcommand{\Zs}{\mathcal{Z}}
\newcommand{\Ys}{\mathcal{Y}}

\newcommand{\Ps}{\mathcal{P}}
\newcommand{\Fs}{\mathcal{F}}
\newcommand{\Gs}{\mathcal{G}}
\newcommand{\Hs}{\mathcal{H}}

\usepackage[accepted]{icml2020}

\usepackage{url}            
\usepackage{booktabs}       
\usepackage{amsfonts}       
\usepackage{nicefrac}       
\usepackage{microtype}      
\usepackage[utf8]{inputenc}

\usepackage{graphicx}
\usepackage{framed}
\usepackage{amssymb}
\usepackage{amsfonts}
\usepackage{mathrsfs}
\usepackage{mathtools}
\usepackage{array}
\usepackage{amsthm}
\usepackage{verbatim} 
\usepackage{enumerate}
\usepackage{bbm}
\usepackage{mathtools}
\usepackage{commath}
\usepackage{wrapfig}
\usepackage{amsbsy}
\usepackage{float}
\usepackage{amsmath}
\usepackage{amsthm}
\usepackage{algorithm}
\usepackage{mathtools}

\usepackage{amssymb}
\usepackage{wrapfig}
\usepackage{xcolor}

\newtheorem{thm}{Theorem}
\newtheorem{theorem}[thm]{Theorem}
\newtheorem{lemma}[thm]{Lemma}
\newtheorem{definition}[thm]{Definition}

\icmltitlerunning{Estimating Generalization under Distribution Shifts via Domain-Invariant Representations}

\begin{document}

\twocolumn[
\icmltitle{Estimating Generalization under Distribution Shifts \\ via Domain-Invariant Representations}




\begin{icmlauthorlist}
\icmlauthor{Ching-Yao Chuang}{mit}
\icmlauthor{Antonio Torralba}{mit}
\icmlauthor{Stefanie Jegelka}{mit}
\end{icmlauthorlist}

\icmlaffiliation{mit}{CSAIL, Massachusetts Institute of Technology, Cambridge, MA, USA}

\icmlcorrespondingauthor{Ching-Yao Chuang}{cychuang@mit.edu}

\icmlkeywords{distribution shift, domain adaptation}

\vskip 0.3in
]



\printAffiliationsAndNotice{} 

\begin{abstract}
  When machine learning models are deployed on a test distribution different from the training distribution, they can perform poorly, but overestimate their performance. In this work, we aim to better estimate a model's performance under distribution shift, without supervision. To do so, we use a set of domain-invariant predictors as a proxy for the unknown, true target labels. Since the error of the resulting risk estimate depends on the target risk of the proxy model, we study generalization of domain-invariant representations and show that the complexity of the latent representation has a significant influence on the target risk. Empirically, our approach (1) enables self-tuning of domain adaptation models, and (2) accurately estimates the target error of given models under distribution shift. Other applications include model selection, deciding early stopping and error detection.
%
\end{abstract}

\section{Introduction}

In many applications, machine learning models are deployed on data whose distribution is different from that of the training data. For instance, self-driving cars must be able to adapt to different weather, change of landscape or traffic, i.e., conditions can change at prediction time. But often, collecting large-scale supervised data on the shifted prediction domain is prohibitively expensive or impossible. While we may hope that the model generalizes to this new data distribution, \emph{estimating empirically} how well a given model will actually generalize is challenging without labels.


Indeed, estimating the \emph{adaptability}, i.e., the generalization to the target distribution, and the related potentially uncertain behavior of a prediction model, is a key concern for AI Safety \citep{amodei2016concrete}, motivating recent work on estimating target performance \citep{steinhardt2016unsupervised, platanios2014estimating}.



In this work, we develop a new idea for estimating performance under distribution shift, by drawing connections with domain adaptation. Necessarily, any method for estimating target performance must make some assumptions. Our method 
assumes the existence of a domain adaptation model that generalizes well from source (training) to target (test). Given the empirical success of domain adaptation, this assumption is met in many practical settings. A prominent class of domain adaptation models, \emph{domain-invariant representations (DIR)} \citep{ben2007analysis,long2015learning,ganin2016domain}, learns a latent, joint representation of source and target data, and a predictor from the latent space to the output labels. In particular, we use a set of ``check" DIR models as a proxy for the unknown, true target labels. If there exist ``good'' domain adaptation models, i.e., they achieve low source and presumably low target error, and those models disagree with the model $h$ we want to evaluate, then the target risk of $h$ is potentially high, and we should not trust it. Our experiments show that this leads to accurate estimates of target error that outperform previous methods.

This idea relies on good domain adaptation models, i.e., our check models should predict well on the target distribution, and not disagree too much with each other.
But, evaluating a domain adaptation model itself on the target distribution is an unsolved problem. Hence, we begin by studying the target error of DIR. We observe that in general, DIR is much more sensitive to model complexity than supervised learning on the source distribution. In particular, the complexity of the representation encoder is key for target generalization and for selecting the set of check models, and points to an important model selection problem. For deep neural networks, this model selection problem essentially means how to optimally divide the network into an encoder and predictor part. Yet, this model selection ideally demands an estimate of target generalization, which we are developing.

We show that, with our framework for estimating target error, it is possible to let DIR models self-tune to find the optimal model complexity. The resulting models achieve good target generalization, and estimate target error of other models well. Our approach applies to estimating the target error of a single or a class of models, and to predicting point-wise error. Hence, it can be used, e.g., for judging reliability and for model selection.
Empirically, we examine our theory and algorithms on sentiment analysis (Amazon review dataset), digit classification (MNIST, MNIST-M, SVHN) and general object classification (Office-31). In short, this work makes the following contributions:
\begin{itemize}\setlength{\itemsep}{-1pt}
\item We develop a generic method for estimating the error of a given model on a new data distribution.
\item We show, theoretically and empirically, the important role of embedding complexity for domain-invariant representations.
\item Our empirical results reflect our analyses and show that the proposed methods work well in practice.
\end{itemize}

\section{Related Work}

\textbf{Estimating risk with distribution shifts.} Estimating model risk on distributions different from the training distribution is important, but difficult with unlabeled data. \citet{platanios2014estimating} construct multiple models based on different views of the data and estimate the risk by calculating agreement rates across models. \citet{steinhardt2016unsupervised} estimate the model's error on distributions very different from the training distribution by assuming a conditional independence structure of the data. \citet{platanios2017estimating} use logical constraints on the data to estimate classification accuracy. Recently, \citet{elsahar2019annotate} evaluate both confidence score and $\mathcal{H}\Delta\mathcal{H}$-divergence to predict performance drop under domain shift. We compare to those methods in the experiments. Different from previous works, we leverage domain-invariant classifiers as proxy target labels. Our method is general in the sense that it can predict the target risk for both domain adaptation and general supervised models.

\textbf{Domain-invariant representations.} DIRs are learned by minimizing a divergence between the embedding of source and target data, and existing approaches for learning DIRs differ in the divergence measure they use. Examples include domain adversarial learning \citep{ganin2014unsupervised, tzeng2015simultaneous, ganin2016domain}, maximum mean discrepancy (MMD) \citep{long2014transfer, long2015learning, long2016unsupervised} and Wasserstein distance \citep{courty2016optimal, courty2017joint, shen2017wasserstein, lee2018minimax}.

Several theoretical frameworks have been proposed to analyze domain-invariant representations. One approach is to bound the target risk by assuming source and target domain share common support.
\citet{wu2019domain} show that exact matching of source and target distributions can increase target risk if label distributions differ between source and target. \citet{johansson2019support} propose generalization bounds based on the overlap of the supports of source and target distribution. However, the assumption of common support fails in most standard benchmarks for domain adaptation. 
Another line of work leverages the $\mathcal{H}\Delta\mathcal{H}$-divergence proposed by \citet{ben2007analysis}. \citet{shu2018dirt} points out that learning domain-invariant representations with disjoint supports can still achieve maximal $\mathcal{H}\Delta\mathcal{H}$-divergence. Recently, \citet{zhao2019learning} 
establish lower and upper bounds on the risk when label distributions between source and target domains differ.  

\section{Unsupervised Domain Adaptation}
\label{sec_uda}

For simplicity of exposition, we consider binary classification with input space $\mathcal X \subseteq \mathbb{R}^{n}$ and output space $\mathcal Y = \{0, 1\}$. The learning algorithm obtains two datasets: labeled source data $\mathcal X_{S}$ from distribution $p_{S}$, and unlabeled target data $\mathcal X_{T}$ from distribution $p_{T}$. We will use $p_S$ and $p_T$ to denote the joint distribution on data and labels $X,Y$ and the marginals, i.e., $p_S(X)$ and $p_S(Y)$. Unsupervised domain adaptation seeks a hypothesis $h: \mathcal X \to \mathcal Y$ in a hypothesis class $\mathcal H$ that minimizes the risk in the target domain measured by a loss function $\ell$ (here, zero-one loss): 
\begin{align}
R_{T}(h) = \mathbb{E}_{x,y \sim p_{T}}[\ell(h(x), y)].
\label{eq:rt}
\end{align}
We do not assume common support in source and target domain. 

\subsection{Domain-invariant Representations}
A common approach to domain adaptation is to learn a joint embedding $g: \mathcal X \to \Zs$ of source and target data \citep{ganin2016domain,tzeng2017adversarial}. The idea is that aligning source and target distributions in a latent space $\mathcal Z$ results in a domain-invariant representation, and hence a subsequent classifier $f: \Zs \to \Ys$ will generalize from source to target. Formally, this results in the following objective function on the hypothesis $h = fg := f \circ g$, where 
we minimize a divergence $d$ between the distributions 
$p_{S}^{g}(Z), p_T^{g}(Z)$ of source and target after the mapping $Z = g(X) \in \mathcal{Z}$:
\begin{align}
\min_{f \in \mathcal{F}, g \in \mathcal{G}} R_{S}(fg) + \alpha d(p_{S}^{g}(Z), p_{T}^{g}(Z)).
\label{di_obj}
\end{align}
The divergence $d$ could be, e.g., the Jensen-Shannon \citep{ganin2016domain} or Wasserstein distance \citep{shen2017wasserstein}. In this paper, we denote the hypothesis class of the entire model $h$ as $\mathcal{H}$, the class of embeddings by $\mathcal{G}$, and the class of predictors by $\mathcal{F}$.

\subsection{Upper Bounds on the Target Risk}

\citet{ben2007analysis} introduced the $\mathcal{H} \Delta \mathcal{H}$-divergence to bound the worst-case loss from extrapolating between domains. Let $R_{D}(h, h^{\prime}) = \mathbb{E}_{x \sim D}[\ell(h(x), h^{\prime}(x))]$ be the
expected disagreement between two hypotheses and an extension of the notation $R_{D}(h) = R_{T}(h, h_{\textnormal{true}})$, where $h_{\textnormal{true}}$ are the true labels. The $\mathcal{H} \Delta \mathcal{H}$-divergence measures whether there is any pair of hypotheses whose disagreement (risk) differs a lot between source and target distribution.

\begin{definition}
\textnormal{($\mathcal{H}\Delta \mathcal{H}$-divergence)} Given two domain distributions $p_{S}$ and $p_{T}$ over $\mathcal{X}$, and a hypothesis class $\mathcal{H}$, the $\mathcal{H}\Delta \mathcal{H}$-divergence between $p_{S}$ and $p_{T}$ is
\begin{align*}
d_{\mathcal{H}\Delta \mathcal{H}}(p_{S}, p_{T}) = \sup_{h, h^{\prime} \in \mathcal{H}} |R_{S}(h, h^{\prime}) - R_{T}(h, h^{\prime})|.
\end{align*}
\end{definition}
The $\mathcal{H}\Delta \mathcal{H}$-divergence is determined by the discrepancy between source and target distribution and the complexity of the hypothesis class $\mathcal{H}$. This divergence allows us to bound the target risk:

\begin{theorem}\label{thm:bendavid}
  \textnormal{\citep{ben2010theory}} 
  For all hypotheses $h \in \mathcal H$, the target risk is bounded as
\begin{align}
R_{T}(h) \leq R_{S}(h) + d_{\mathcal{H}\Delta \mathcal{H}}(p_{S}, p_{T}) + \lambda_{\mathcal H},
\label{bound_ben}
\end{align}
where $\lambda_{\mathcal H}$ is the best joint risk 
\begin{align*}
\lambda_{\mathcal H} \coloneqq \inf_{h' \in \mathcal{H}}[R_S(h') + R_T(h')].
\end{align*}
\end{theorem} 

Similar results exist for continuous labels \citep{cortes2011domain,mansour2009domain}. Theorem \ref{thm:bendavid} has been an influential theoretical result in domain adaptation, and motivated work on domain invariant representations. For example, recent work (\citet{ganin2016domain,johansson2019support}) applied Theorem \ref{thm:bendavid} to the hypothesis class $\mathcal F$ that maps the representation space $\mathcal{Z}$ induced by an encoder $g$ to the output space:
\begin{align}
R_{T}(fg) \leq R_{S}(fg) + d_{\mathcal F \Delta \mathcal F}(p_{S}^{g}(Z), p_{T}^{g}(Z)) + \lambda_{\mathcal F}(g)
\label{bound1}
\end{align}
where $\lambda_{\mathcal F}(g)$ is the best hypothesis risk with fixed $g$, i.e., $\lambda_{\mathcal F}(g) \coloneqq \inf_{f' \in \mathcal{F}}[R_S(f'g) + R_T(f'g)]$. The $\mathcal F \Delta \mathcal F$ divergence implicitly depends on the fixed $g$ and can be small if $g$ provides a suitable representation. However, if $g$ induces a wrong alignment, then the best hypothesis risk $\lambda_{\mathcal F}(g)$ is large with any function class $\mathcal{F}$. 

\section{Estimating Target Risk: Main Idea}
\label{sec_idea}

\begin{figure}[]
\begin{center}   
\includegraphics[width=0.8\linewidth]{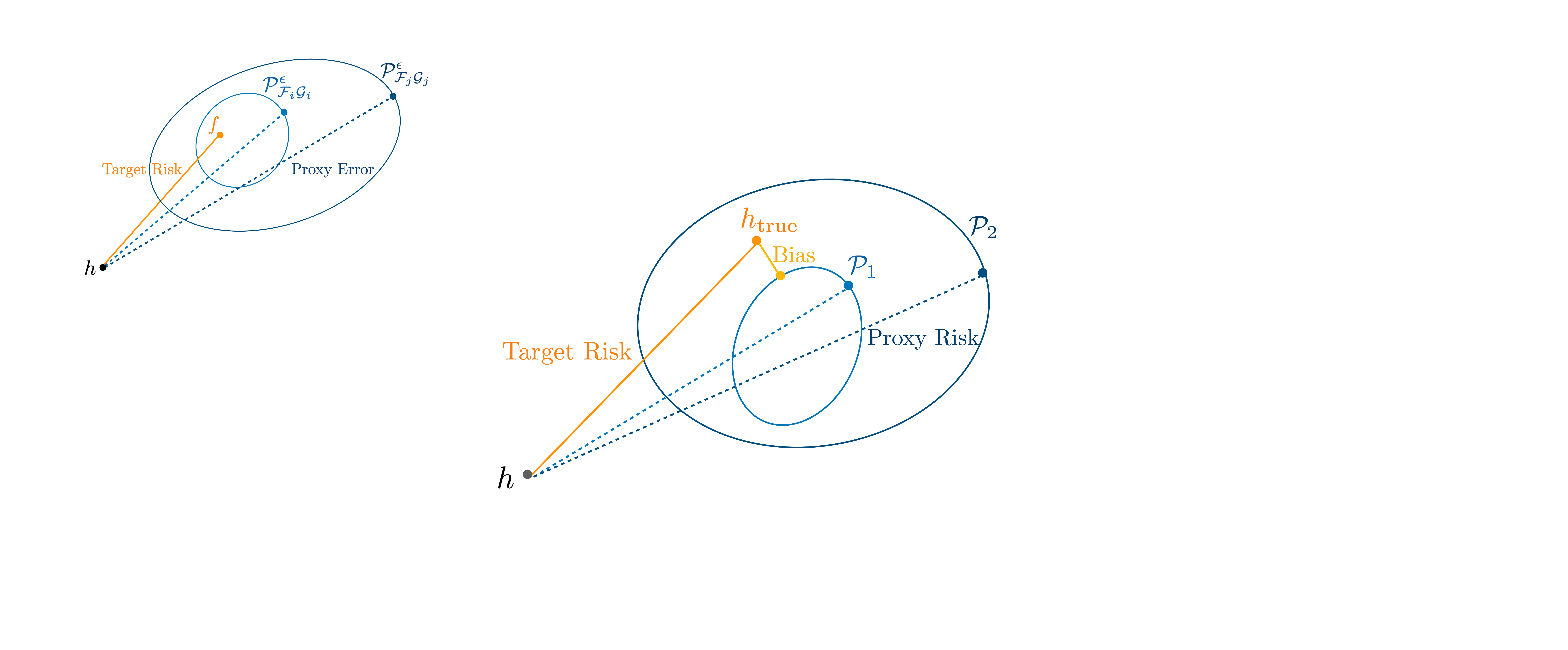}
\end{center}
\caption{\textcolor{black}{Conceptual illustration of proxy risk: the orange line is the true target risk, the dashed lines are the proxy risks with respect to two sets of check models, $\mathcal{P}_1$ and $\mathcal{P}_2$ where $\mathcal{P}_1 \subseteq \mathcal{P}_2$. By construction, although $h_\textnormal{true} \in \mathcal{P}_2$ (zero bias), the proxy risk calculated with $\mathcal{P}_2$ is not tight enough to approximate the target risk well. In contrast, $\mathcal{P}_1$ has a nonzero bias (yellow line) but tighter estimation.}} \label{fig_method}
\end{figure}

Our goal is to estimate the error of a given, learned model $h$ on a target distribution $p_T$, without observing true labels on the target. Let $h_{\text{true}}$ be the true labeling, and $h^* = \arg\inf_{h \in \Ps}R_T(h)$. By the triangle inequality, $R_T(h) = R_T(h,h_{\text{true}}) \leq R_T(h,h^*) + R_T(h^*)$. The main idea underlying our approach is to obtain an upper bound on $R_T(h)$ by replacing $h^*$ with candidates from a set of proxy models $\Ps$ that we also call \emph{check models}. 
\begin{lemma}\label{lemma_bv}
Given a hypothesis class $\mathcal{P}$, for all $h \in \mathcal{H}$,
\begin{align}
R_{T}(h) 
&\leq \underbrace{\sup_{h^{\prime } \in \mathcal{P}}R_{T}(h, h^{\prime})}_{\textnormal{Proxy Risk}} + \underbrace{\inf_{h^{\prime} \in \mathcal{P}}R_{T}(h^{\prime})}_{\textnormal{Bias}}.
\label{bound5}
\end{align}

\end{lemma}
We prove all theoretical results in the Appendix A. The first term in Lemma~\ref{lemma_bv} measures the maximal disagreement (risk) between the hypothesis $h$ and a check model $h^{\prime} \in \mathcal{P}$, instead of $h^*$. The second term measures how good the check models are. For this bound to be tight, $\Ps$ must contain a good hypothesis. At the same time, $\Ps$ should not contain any unnecessarily disagreeing hypotheses, otherwise the proxy risk will be too large. Figure \ref{fig_method} provide an illustrative example of the idea.

\paragraph{Connection to Domain Adaptation}

The proxy risk can be estimated empirically. If the bias term is small, namely, there exists a good hypothesis in the check models, then the proxy risk itself is a good estimate of an upper bound on $R_T(h)$. It remains to determine the set $\Ps$. 

\begin{lemma}\label{lemma_err}
Given a hypothesis class $\mathcal{P}$, for all $h \in \mathcal{H}$,
\begin{align}
 \underbrace{|\sup_{h^{\prime} \in \mathcal{P}} R_{T}(h, h^{\prime}) - R_{T}(h)|}_{\textnormal{Estimation Error}} \leq \sup_{h^{\prime} \in \mathcal{P}}R_{T}(h^{\prime}).
\end{align}

\end{lemma}
\textcolor{black}{
Lemma~\ref{lemma_err} links our approach with domain adaptation: the target risk of the check models affects the error of estimating risk via the proxy risk. This motivates domain adaptation models as check models, because they are designed to minimize the target risk. In Section~\ref{sec_risk_est}, where we develop this idea in detail, $\Ps$ is the set of all DIR models that have low DIR objective. To understand the tightness of our proxy risk-based estimation, we begin with a closer look at what affects the target risk of domain invariant representations.}

\section{Understanding the Adaptability of DIR}
\label{sec:da_bounds}
In this section, we aim to better understand what affects target risk and ambiguity on the target for domain invariant representations. The bound~\eqreff{bound1} highlights the effect of the complexity of the prediction models $\Fs$, and of the quality of the alignment via the embedding $g$. But, as the following toy example illustrates, another important component is the complexity of the embedding class $\Gs$.


\begin{figure}[]
\begin{center}   
\includegraphics[width=0.9\linewidth]{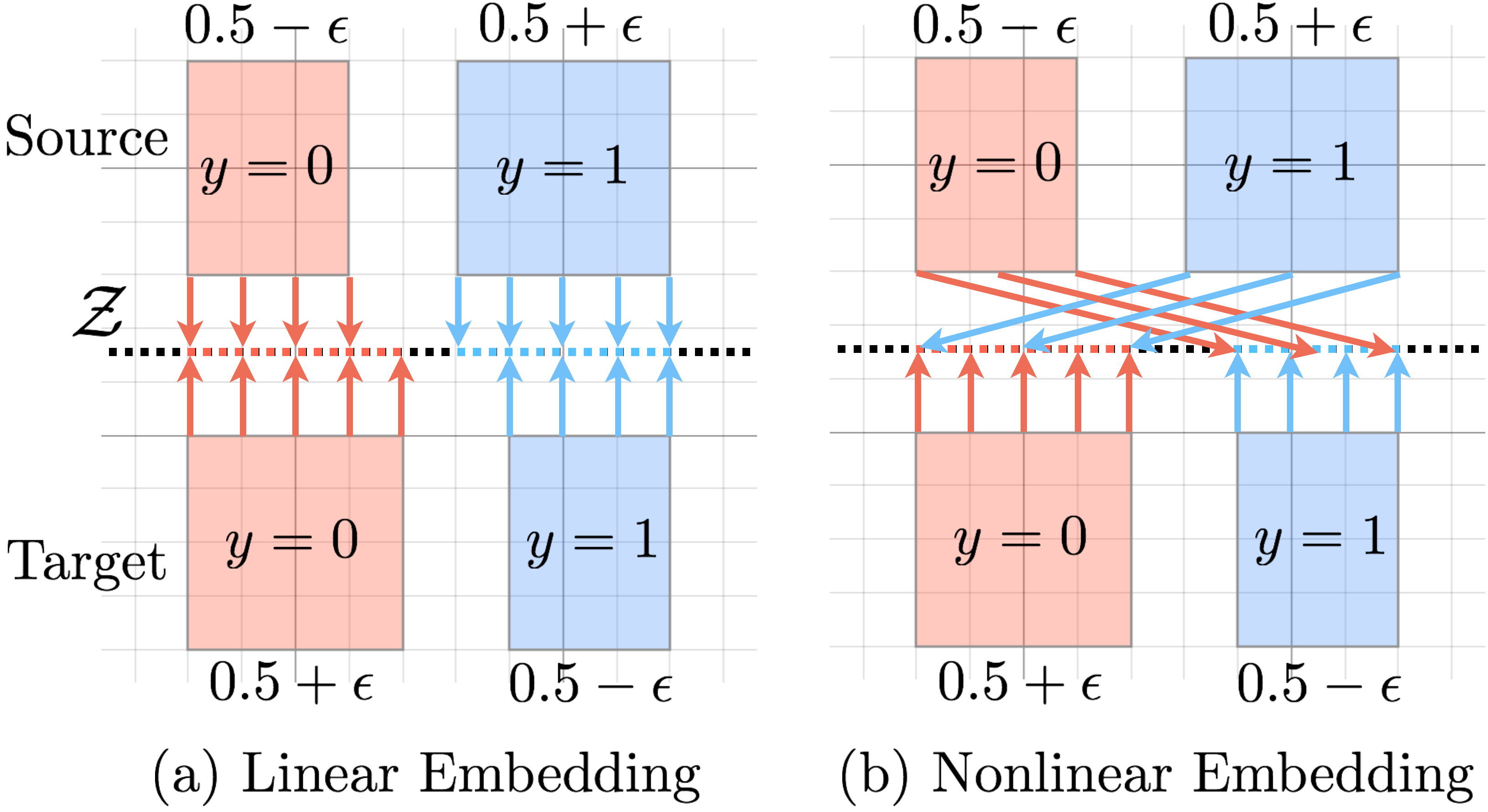}
\end{center}
\caption{Illustrative example in 2D. The 1D representation space is illustrated as a dotted line, and arrows indicate the embedding from 2D to 1D. (a) Optimal embedding when $\mathcal{G}$ is the class of linear functions. 
    (b) Optimal embedding with a complex nonlinear encoder class.} \label{fig_2d}
\end{figure}

\paragraph{Toy Example.}
Figure~\ref{fig_2d} shows a binary classification problem in 2D with disjoint support and a slight shift in the label distributions from source to target: $p_{S}(y=1) = p_{T}(y=1) + 2\epsilon$. For a 1D latent representation space, if we allow arbitrary maps $g \in \Gs$, then, e.g., a complicated nonlinear map as in Figure~\ref{fig_2d}(b) can achieve zero DIR objective value (\eqref{di_obj}), but maximum target risk $R_T(fg) = 1$. If we restrict $\Gs$ to linear maps, then a map $g$ as in Figure~\ref{fig_2d}(a) achieves optimal DIR objective value of $2\epsilon$, and minimum target risk.
Hence, a too powerful embedding class $\Gs$ can increase ambiguity, variance and hence target risk.

\begin{figure}[]
\begin{center}   
\includegraphics[width=\linewidth]{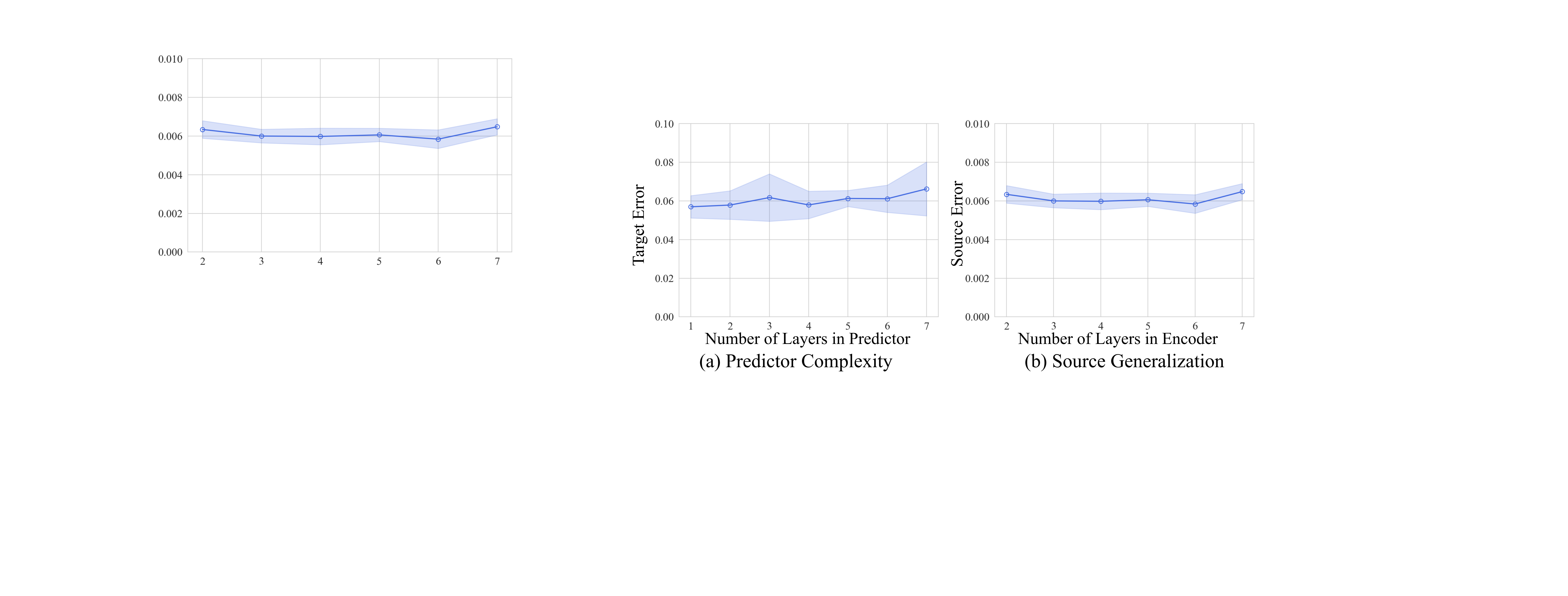}
\end{center}
\caption{(a) Effect of predictor complexity on target generalization for MNIST$\rightarrow$MNIST-M and (b) effect of embedding complexity on source (MNIST) generalization.} \label{fig_source}
\end{figure}

\paragraph{Empirical Effect of Complexity.} In the experiments in Section~\ref{exp}, e.g., in Figure~\ref{fig_amazon_office_1}, we observe that throughout, the complexity of $\Gs$ has a noticeable effect on the target risk. In contrast, in analogous experiments for the predictor $f$ shown in Figure~\ref{fig_source}(a), the predictor class $\Fs$ has a much weaker influence. Likewise, Figure~\ref{fig_source}(b) demonstrates that generalization on the source domain, i.e., ``normal'' generalization of supervised learning, is also much less affected by the model complexity. In summary, empirically, the adaptability of domain-invariant representations is more sensitive to model complexity than supervised learning, and in general most sensitive to the complexity of the embedding class $\Gs$. Hence, we focus on embedding complexity.

\subsection{Bounds for Domain-invariant Representations}
Motivated by the above observations, we next expose how the bound on the target risk depends on the complexity of the embedding class. Directly applying Theorem~\ref{thm:bendavid} to the composition $\mathcal{H} = \mathcal{F}\mathcal{G}$ treats both jointly and does not make the role of the embedding very explicit. Instead, we define a version of the $\mathcal{H}\Delta\mathcal{H}$-divergence that explicitly measures variation of the embeddings in $\mathcal{G}$:
\begin{definition}
\textnormal{($\mathcal{F}_{\mathcal{G}\Delta \mathcal{G}}$-divergence)} For two domain distributions $p_{S}$ and $p_{T}$ over $\mathcal{X}$, an encoder class $\mathcal{G}$, and predictor class $\mathcal{F}$, the $\mathcal{F}_{\mathcal{G}\Delta \mathcal{G}}$-divergence between $p_{S}$ and $p_{T}$ is
\begin{align*}
      d_{\mathcal{F}_{\mathcal{G}\Delta\mathcal{G}}}(p_{S}, p_{T}) = \sup_{f \in \mathcal{F}; \; g,g^{\prime} \in \mathcal{G}} |R_{S}(fg, fg^{\prime}) - R_{T}(fg, fg^{\prime})|.
\end{align*}
\label{ordef}
\vspace{-4mm}
\end{definition}  
Importantly, the $\mathcal{F}_{\mathcal{G}\Delta \mathcal{G}}$-divergence is smaller than the $(\mathcal{FG})\Delta (\mathcal{FG})$-divergence, since the two hypotheses in the supremum, $fg$ and $fg^{\prime}$, share the same predictor $f$. 
\begin{theorem}\label{thm:main}
  For all $f \in \mathcal F$ and $g \in \mathcal G$,
\begin{align}
R_{T}(fg) \leq R_{S}(fg) &+ \underbrace{d_{\mathcal{F}\Delta\mathcal{F}}(p_{S}^{g}(Z), p_{T}^{g}(Z))}_{\text{\textnormal{Latent Divergence}}}  \nonumber\\ &+ \underbrace{d_{\mathcal{F}_{\mathcal{G}\Delta\mathcal{G}}}(p_{S}, p_{T})}_{\substack{\textnormal{Embedding Complexity}}}+\lambda_{\mathcal F\mathcal{G}}(g). \label{bound4}
\end{align}
where $\lambda_{\mathcal F\mathcal{G}}(g)$ is a variant of the best in-class joint risk:
\begin{align}
    \lambda_{\mathcal F\mathcal{G}}(g) =  \inf_{f^{\prime} \in \mathcal{F}, g^{\prime} \in \mathcal{G}} 2R_{S}(f^{\prime}g) + R_{S}(f^{\prime}g^{\prime}) + R_{T}(f^{\prime}g^{\prime}). \nonumber
\end{align}
\vspace{-4.5mm}
\end{theorem}

This target generalization bound is small if (C1) the source risk is small, (C2) the latent divergence is small, because the domains are well-aligned and/or $\mathcal{F}$ is restricted, (C3) the complexity of $\mathcal{G}$ is restricted to avoid overfitting of alignments, and (C4) good source and target risk is in general achievable with $\mathcal{F}$ and $\mathcal{G}$ and the encoder is good for the source domain. The bound naturally explains the tradeoff we observe in the subsequent experiments between the following terms: the latent divergence (which increases with complexity of $\mathcal{F}$ and decreases with complexity of $\mathcal{G}$), embedding complexity (which increases with complexity of $\mathcal{F}$ and $\mathcal{G}$), and the best in-class joint risk (which decreases with complexity of $\mathcal{F}$ and $\mathcal{G}$). Overly expressive encoders suffer from a larger embedding complexity penalty, while smaller encoders fail to minimize the latent divergence.

\subsection{Experiments}
\label{exp}

\begin{figure}[]
\begin{center}   
\includegraphics[width=\linewidth]{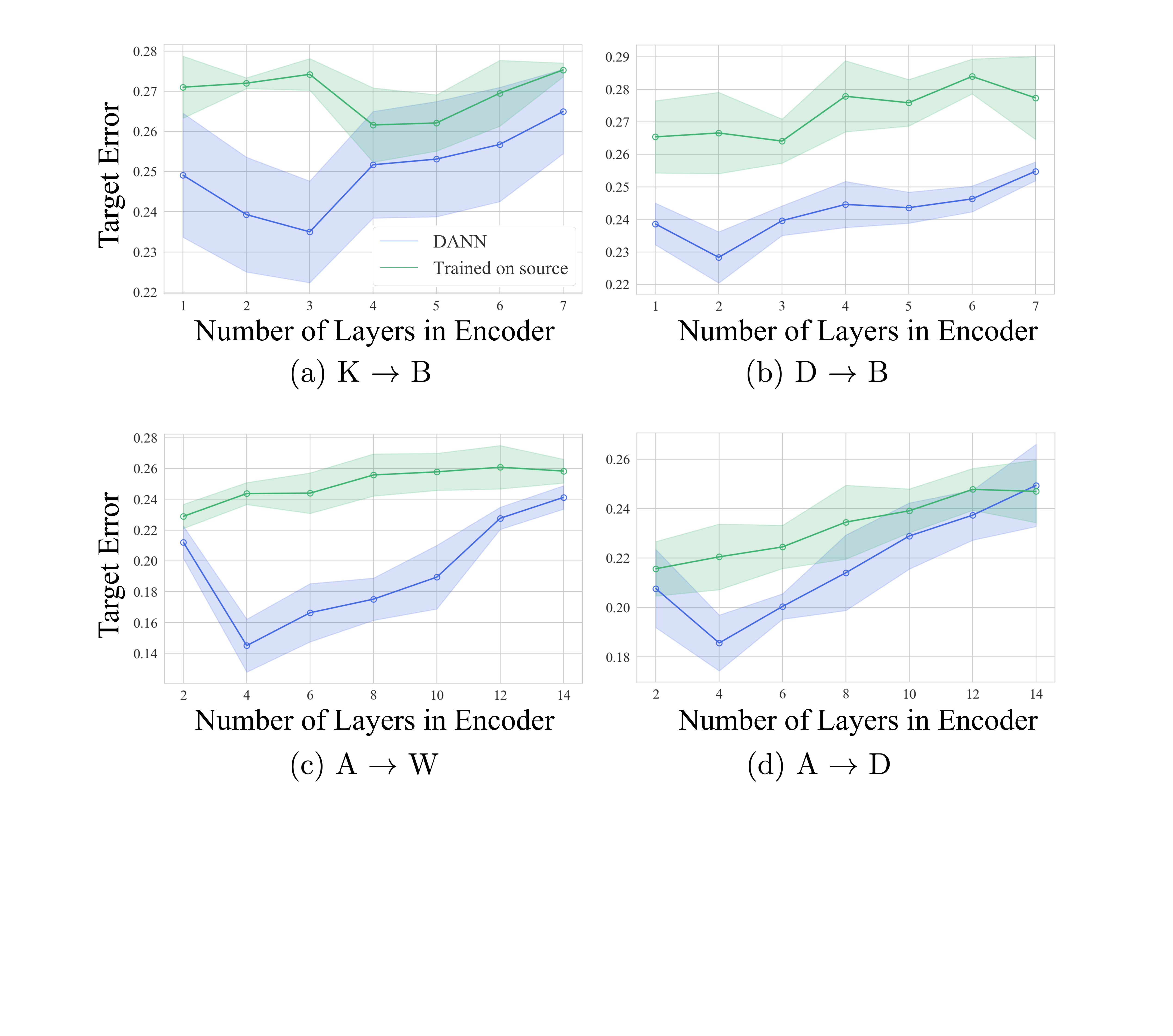}
\end{center}
\caption{Effect of embedding complexity on target risk.  First row: Sentiment Classification. Second row: Object Classification.} \label{fig_amazon_office_1}
\end{figure}

Next, we probe the effect of embedding complexity empirically, via experiments with several standard benchmarks: sentiment analysis (Amazon reviews dataset), digit classification (MNIST, MNIST-M, SVHN) and general object classification (Office-31). In all experiments, we train DANN \citep{ganin2016domain}, which measures the latent divergence via a domain discriminator (Jensen Shannon divergence).  A validation set from the source domain is used as an early stopping criterion during learning. In all experiments, we use a progressive training strategy for the discriminator \citep{ganin2016domain}. We primarily consider two types of complexity: number of layers and number of hidden neurons. In all embedding complexity experiments, we retrain each model for 5 times and plot the mean and standard deviation of the target error. Dataset and architecture details may be found in the appendix.

\paragraph{Sentiment Classification.}
We first examine complexity tradeoffs on the Amazon reviews data, which has four domains: books (B), DVD disks (D), electronics (E), and kitchen appliances (K). The hypothesis class are multi-layer ReLU networks. We show results for K$\rightarrow$B, and D$\rightarrow$B in Figure~\ref{fig_amazon_office_1} and defer the rest to the appendix. To probe the effect of embedding complexity, we fix the predictor class to $4$ layers and vary the number of layers of the embedding. Figure~\ref{fig_amazon_office_1} shows that the target error decreases initially, and then increases as more layers are added to the encoder. 

\paragraph{Object Classification.}
Office-31 \citep{saenko2010adapting}, one of the most widely used benchmarks in domain adaptation, contains three domains: Amazon (A), Webcam (W), and DSLR (D) with 4,652 images and 31 categories. We show results for A$\rightarrow$W, A$\rightarrow$D in Figure~\ref{fig_amazon_office_1}, and the rest in the Appendix B. To overcome the lack of training data, similar to \citep{li2018domain, long2018conditional}, we use ResNet-50 \citep{he2016deep} pretrained on ImageNet \citep{deng2009imagenet} for feature extraction. With the extracted features, we adopt multi-layer ReLU networks as hypothesis class. Again, we increase the depth of the encoder while fixing the depth of the predictor to $2$. Even with a powerful feature extractor, the embedding complexity tradeoff still exists.

\begin{figure}[]
\begin{center}   
\includegraphics[width=\linewidth]{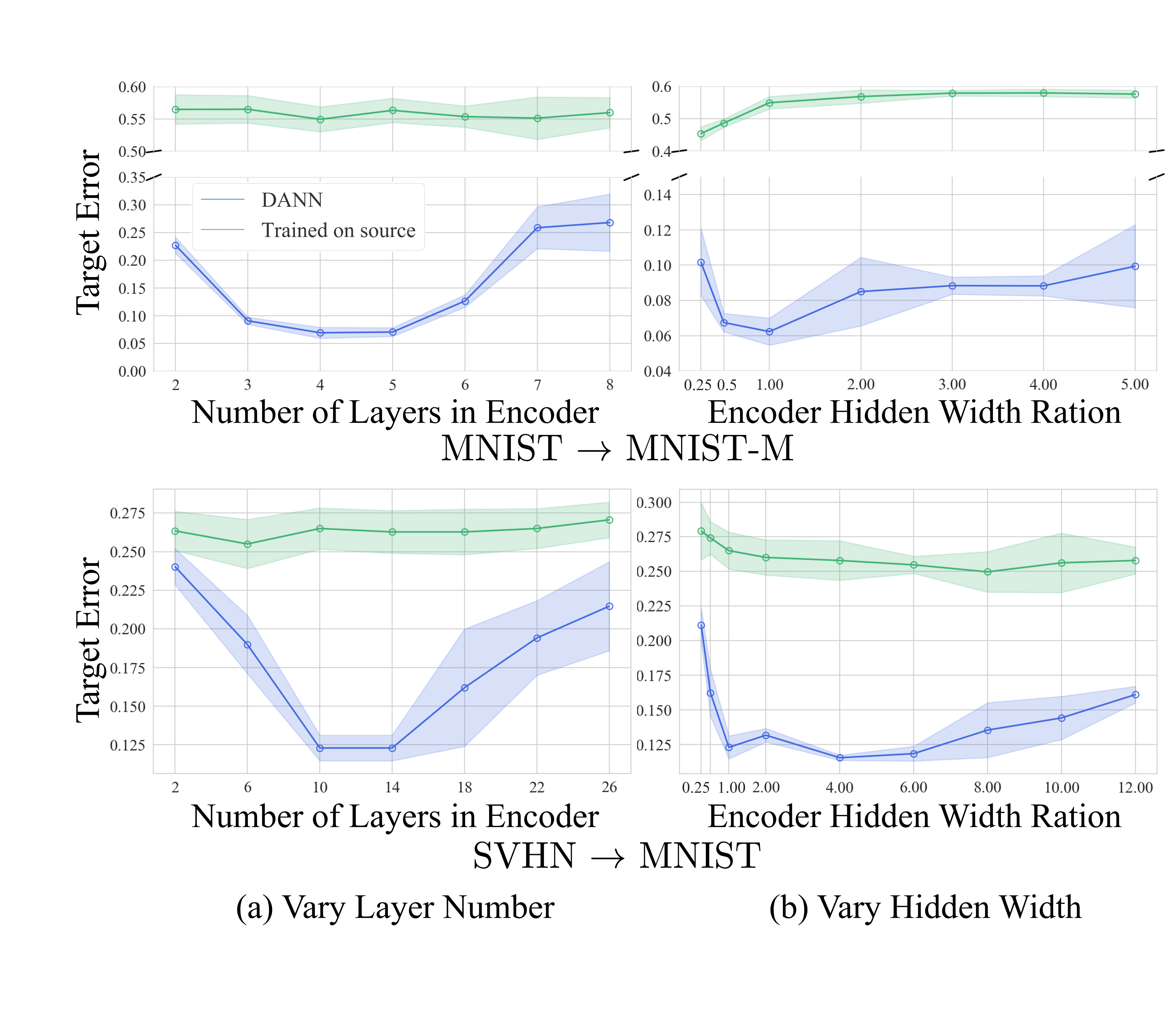}
\end{center}
\caption{Effect of embedding complexity on target risk for digit classification.} \label{fig_digit_1}
\end{figure}

\paragraph{Digit Classification.}

We next verify our findings on standard  domain adaptation benchmarks: MNIST$\rightarrow$MNIST-M (M$\rightarrow$M-M) and SVHN$\rightarrow$MNIST (S$\rightarrow$M). We use standard CNNs as the hypothesis class.

To analyze the effect of the embedding complexity, we augment the original two-layer CNN encoders with 1 to 6 additional CNN layers for M$\rightarrow$M-M and 1 to 24 for S$\rightarrow$M, leaving other settings unchanged. Figure \ref{fig_digit_1}(a) shows the results. Again, the target error decreases initially and increase as the encoder becomes more complex. Notably, the target error increases by $19.8\%$ in M$\rightarrow$M-M and $8.8\%$ in S$\rightarrow$M compared to the optimal case, when more layers are added to the encoder. We also consider the width of hidden layers as a complexity measure,
while fixing the depth of both encoder and predictor. The results are shown in Figure \ref{fig_digit_1}(b). This time, the decrease in target error is not significant compared to increasing encoder depth. This suggests that depth plays a more important role than width in learning domain-invariant representations. In the appendix, we also investigate the importance of inductive bias and for domain-invariant representations.

\section{Division for Multilayer Neural Networks}
\vspace{-1mm}
\label{sec_bo}
Next, we adapt the bound in Theorem \ref{thm:main} to multilayer networks. Specifically, we consider the number of layers as a measure of complexity. Assume $\mathcal{H}$ is the class of $N$-layer feedforward neural networks with a fixed width. The model 
%
$h \in \Hs$ can be decomposed as $h = f_{i}g_{i} \in \mathcal{F}_{i}\mathcal{G}_{i} = \mathcal{H}$ for $i \in \{1, 2, \dots, N-1 \}$, where the embedding $g_{i}$ is formed by the first layer to the $i$-th layer and the predictor $f_{i}$ is formed by the ($i+1$)-th layer to the last layer. We can then rewrite the bound in Theorem \ref{thm:main} in layer-specific form:
\begin{align}
\label{eq_lsb}
R_{T}(h) \leq &\, R_{S}(h) + \underbrace{d_{\mathcal{F}_{i}\Delta\mathcal{F}_{i}}(p_{S}^{g_{i}}(Z), p_{T}^{g_{i}}(Z))}_{\substack{\text{\textnormal{Latent Divergence}}  \text{\textnormal{ in $i$-th layer}} }} \nonumber\\&+ \underbrace{d_{{\mathcal{F}_{i}}_{\mathcal{G}_{i}\Delta\mathcal{G}_{i}}}(p_{S}, p_{T})}_{\substack{\text{\textnormal{Embedding Complexity}}  \text{\textnormal{ w.r.t $\mathcal{G}_i$}} }}+\lambda_{\mathcal F_{i}\mathcal{G}_{i}}(g_{i}).
\end{align}
Minimizing the domain-invariant loss in different layers leads to different tradeoffs between fit and complexity penalties. This is reflected by the following inequalities that relate different layer divisions.
\newtheorem{proposition}[thm]{Proposition}
\begin{proposition}\label{prop}
In an $N$-layer feedforward neural network $h = f_{i}g_{i} \in \mathcal{F}_{i}\mathcal{G}_{i} = \mathcal{H}$, the following inequalities hold for all $i \leq j \leq N-1$:
\begin{align}
    d_{{\mathcal{F}_{i}}_{\mathcal{G}_{i}\Delta\mathcal{G}_{i}}}(p_{S}, p_{T}) &\leq
    d_{{\mathcal{F}_{j}}_{\mathcal{G}_{j}\Delta\mathcal{G}_{j}}}(p_{S}, p_{T})   \nonumber\\
         d_{{\mathcal{F}_{i}}\Delta\mathcal{F}_{i}}(p_{S}^{g_{i}}(Z), p_{T}^{g_{i}}(Z)) &\geq
    d_{{\mathcal{F}_{j}}\Delta\mathcal{F}_{j}}(p_{S}^{g_{j}}(Z), p_{T}^{g_{j}}(Z)).  \nonumber
\end{align}
\end{proposition} 
Proposition \ref{prop} states that a deeper embedding allows for better alignments and simultaneously reduces the depth (power) of $\mathcal{F}$; both reduce the latent divergence. At the same time, it incurs a larger $\mathcal{F}_{\mathcal{G}\Delta\mathcal{G}}$-divergence. This is a tradeoff within the fixed combined hypothesis class $\mathcal{H}$.

This suggests that there might be an optimal division that minimizes the bound on the target risk. In practice, this translates into the question: \textit{in which intermediate layer should we optimize the domain-invariant loss?}

\subsection{Experiments}
\label{exp_2}
\begin{figure}[]
\begin{center}   
\includegraphics[width=\linewidth]{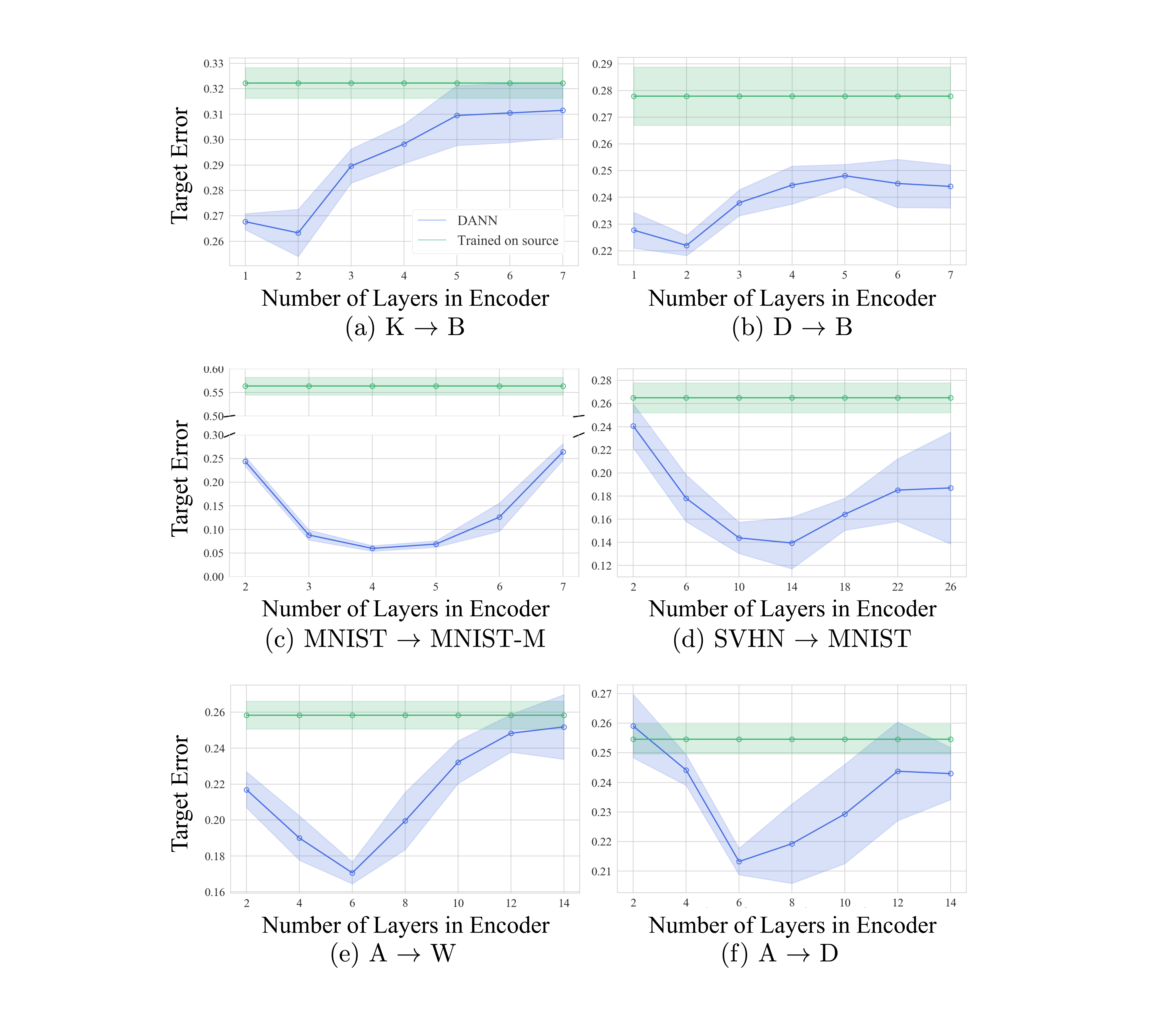}
\end{center}
\vspace{-3mm}
\caption{The effect of layer division in fixed-depth neural networks. First row: Sentiment Classification; Second row: Digit Classification; Third row: Object Classification.} \label{fig_fg}
\vspace{-3mm}
\end{figure}

Next, we examine the embedding complexity tradeoff when the total number of layers is fixed, with the setup of Section \ref{exp}. We first probe the tradeoff when the total number of layers is fixed to $8$ for sentiment classification. The results in Figure~\ref{fig_fg} suggest that there exists an optimal setting for all tasks. Next, we fix the total number of CNN layers of the neural network to $7$ and $26$ for M$\rightarrow$M-M and S$\rightarrow$M, respectively, and optimize the domain-invariant loss in different intermediate layers. The results again show a ``U-curve", indicating the existence of an optimal division. Even with this fixed total size of the network ($\mathcal{H}$), the performance gap between different divisions can still reach $19.5\%$ in M$\rightarrow$M-M and $10.4\%$ in S$\rightarrow$M. Similar results can be seen in object classification with fixed total network depth 14. More experimental results may be found in the appendix.

In summary, empirically, there is an optimal division with minimum target error, suggesting that for a fixed $\mathcal{H}$, i.e., total network depth, not all divisions are equal. We will provide methods to predict the optimal division in Section~\ref{sec_est_exp}.

\section{Estimating Target Risk}\label{sec_risk_est}

Upper bounds such as the ones in Sections \ref{sec_uda} and \ref{sec:da_bounds} are useful for theoretical insights and intuition about effects, but hard to compute explicitly. Here, we return to the idea in Section~\ref{sec_idea} to estimate the target risk of a given model by using a selected set $\Ps$ of check models as proxies. In particular, given the bound in Lemma~\ref{lemma_bv}, we use $\sup_{h^{\prime } \in \mathcal{P}}R_{T}(h, h^{\prime})$ as an estimate of the target risk. Our approach works well if $\inf_{h^{\prime} \in \mathcal{P}}R_{T}(h^{\prime})$ is small, i.e., if there is a good target prediction model in $\Ps$.

We define the set $\Ps$ of check models to be all domain-invariant classifiers that achieve low DIR objective value, i.e., they achieve low source risk and align the source and target distributions well:
\begin{align}
    &\mathcal{P}_{\mathcal{FG}}^{\epsilon} = \{h=fg \in \mathcal{FG}| R_{S}(h) + \alpha d(p_{S}^{g}(Z), p_{T}^{g}(Z)) \leq \epsilon \}. \nonumber
\end{align}
Hence, we implicitly assume that there exists some DIR model that achieves low target risk.

\subsection{Connection to Embedding Complexity}
How good is the resulting proxy risk as an estimate of the target risk of $h$? Lemma~\ref{lemma_err} states that the target risk of the check models gives an upper bound on the estimation error:
\begin{align}
 |\sup_{h^{\prime} \in \mathcal{P}_{\mathcal{FG}}^{\epsilon}} R_{T}(h, h^{\prime}) - R_{T}(h)| \leq \sup_{h^{\prime} \in \mathcal{P}_{\mathcal{FG}}^{\epsilon}}R_{T}(h^{\prime}).
\end{align}

Recall that the set $\mathcal{P}_{\mathcal{FG}}^{\epsilon}$ comprises all DIR models that achieve low DIR objective value. If $\sup_{h^{\prime} \in \mathcal{P}_{\mathcal{FG}}^{\epsilon}}R_{T}(h^{\prime})$ is large, then the DIR objective is not sufficiently determining to identify a good target classifier, and generalization to the target is impossible. The results in Section~\ref{sec:da_bounds} suggest that the embedding complexity of the 
DIR check models plays an important role for target generalization. To minimize the estimation error, we should select a class of DIR models with suitable embedding complexity, i.e., one with an optimal division. As we will show in Section \ref{sec_exp_optimal_div}, it is possible to also use our ideas to let DIR models self-tune, to find the optimal embedding complexity.

\subsection{Computing the Target Risk Estimator}
\label{sec_approx}
To approximate the proxy risk $\sup_{h^{\prime} \in \mathcal{P}_{\mathcal{FG}}^{\epsilon}} R_{T}(h, h^{\prime})$, we aim to maximize the disagreement under model constraints:
\begin{align}
    \max_{f^{\prime}g^{\prime} \in \mathcal{FG}}& R_{T}(h, f^{\prime}g^{\prime}) \\\text{\;s.t.\;}&  R_{S}(f^{\prime}g^{\prime}) + \alpha  d(p_{S}^{g^{\prime}}(Z), p_{T}^{g^{\prime}}(Z)) \leq \epsilon
\end{align}
Computationally, it is more convenient to replace the constraint with a penalty via Lagrangian relaxation:
%
\begin{align}
    \max_{f^{\prime}g^{\prime} \in \mathcal{FG}} R_{T}(h, f^{\prime}g^{\prime}) - \lambda( R_{S}(f^{\prime}g^{\prime}) + \alpha d(p_{S}^{g^{\prime}}(Z), p_{T}^{g^{\prime}}(Z))
    \nonumber
\end{align}
where $\lambda > 0$. We use empirical estimates for $R_T$, $R_S$, and minimize the empirical objective via standard stochastic gradient descent. 

Algorithm \ref{algo1} provides details about approximating the proxy risk\footnote{The code is available at \url{https://github.com/chingyaoc/estimating-generalization}.}. In brief, we first pretrain $h^{\prime} = f^{\prime}g^{\prime}$, and then maximize the disagreement with $h$ under constraints. Empirically we maximize the disagreement on the training set and check the constrains $R_{S}(f^{\prime}g^{\prime}) + \alpha d(p_{S}^{g^{\prime}}(Z), p_{T}^{g^{\prime}}(Z)) \leq \epsilon$ with the validation set.

\begin{algorithm}
\caption{Computing Proxy Risk}
\begin{algorithmic}
\REQUIRE Target hypothesis $h$; Check model class $\mathcal{H} = \mathcal{FG}$; $S_{S}$ and $S_{T}$: labeled source dataset and unlabeled target dataset; $\alpha, \lambda, \epsilon$: tradeoff parameters; $T_1$: Epochs for training domain-invariant classifier; $T_2$: Epochs for maximizing the disagreement.
\STATE
\STATE $\rhd$ \textit{Pretrain check model} $h^{\prime}$
\STATE Initialize $h^{\prime} = f^{\prime}g^{\prime} \in \mathcal{FG}$
\STATE Train $h^{\prime}$ for $T_{1}$ epochs to minimize $R_{S}(h^{\prime}) + \alpha d(p_{S}^{g^{\prime}}(Z), p_{T}^{g^{\prime}}(Z))$
\STATE
\STATE $\rhd$ \textit{Maximize the disagreement}
\STATE Initialize MaxRisk $ = 0$
\FOR{$i=1,\dots,T_{2}$}
\STATE Train $h^{\prime}$ for one epoch to minimize $-R_{T}(h, h^{\prime}) + \lambda( R_{S}(f^{\prime}g^{\prime}) + \alpha d(p_{S}^{g^{\prime}}(Z), p_{T}^{g^{\prime}}(Z))$
\IF{$R_{S}(f^{\prime}g^{\prime}) + \alpha d(p_{S}^{g^{\prime}}(Z), p_{T}^{g^{\prime}}(Z) \leq \epsilon$ and $R_{T}(h, h^{\prime})$ $\geq$ MaxRisk}
\STATE Set MaxRisk $ = R_{T}(h, h^{\prime}$)
\ENDIF
\ENDFOR
\STATE \textbf{return} MaxRisk
\label{algo1}
\end{algorithmic}
\end{algorithm}

\section{Experiments}
We evaluate our method on two broad tasks: model selection for DIR models and estimating target risk of any given model. Throughout, the experimental settings and the model architectures are the same as in Section~\ref{exp_2}.

\subsection{Model Selection for DIR}
\label{sec_est_exp}

\paragraph{Estimating Optimal Network Division}
\label{sec_exp_optimal_div}

We begin with estimating the optimal layer division of a DIR model into encoder and predictor that minimizes target risk. By Lemma~\ref{lemma_err}, this will yield a good class of check models. To estimate the DIR models' target risk, we follow the same strategy as in Section~\ref{sec_idea}, but for a \emph{class} of models: the worst target error for division $i$ can be bounded with a second level of proxy classifiers:
\begin{align}
    \sup_{h \in \mathcal{P}_{\mathcal{F}_i\mathcal{G}_i}^{\epsilon}}R_{T}(h) \leq \underbrace{\sup_{h \in \mathcal{P}_{\mathcal{F}_i\mathcal{G}_i}^{\epsilon} \atop h^{\prime } \in \mathcal{P}_{\mathcal{F'}\mathcal{G'}}^{\epsilon}}R_{T}(h, h^{\prime})}_{\textnormal{Worst In-class Proxy Risk}} + \inf_{h^{\prime} \in \mathcal{P}_{\mathcal{F}\mathcal{G}}^{\epsilon}}R_{T}(h^{\prime}). \nonumber
\end{align}
We select the division that minimizes the worst in-class proxy risk.

\begin{figure}[]
\begin{center}   
\includegraphics[width=\linewidth]{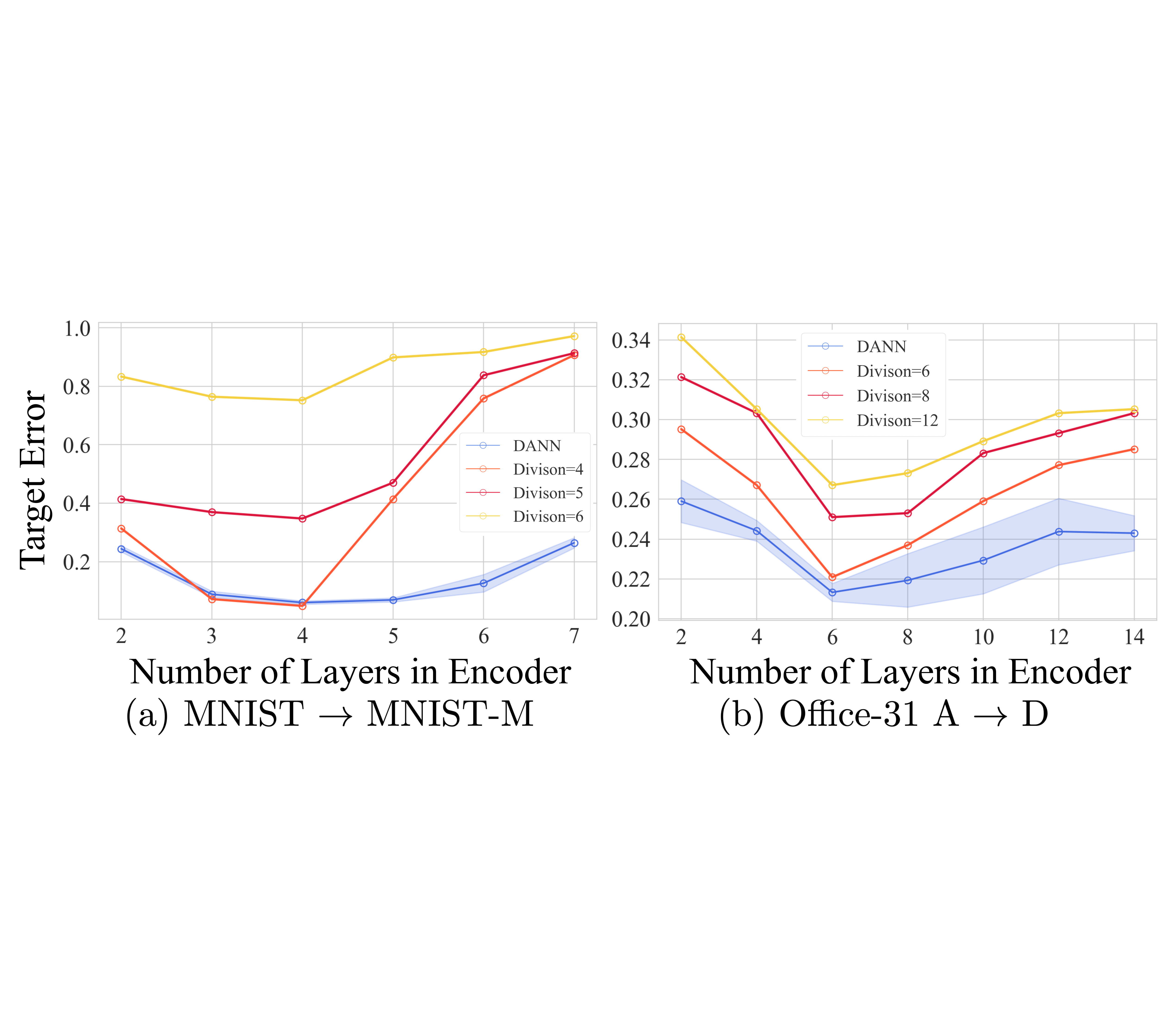}
\end{center}
\caption{Estimating optimal network division via in-class proxy risk. Different colors indicate second-level check models with different divisions. ``DANN'' is the true target test error.} \label{fig_od}
\end{figure}

Computationally, we adopt the approach from Section~\ref{sec_approx} to approximate the worst in-class proxy risks. Figure \ref{fig_od} shows the true target test error for a DIR model, computed with labels (blue line), for different divisions, compared to our in-class proxy risk estimates. The different lines correspond to different second-level check models. The results suggest that (1) we can accurately estimate the best layer division \emph{without supervision}, and (2) this self-tuning strategy is robust to the choice of second-level check models.

\begin{figure}[]
\begin{center}   
\includegraphics[width=\linewidth]{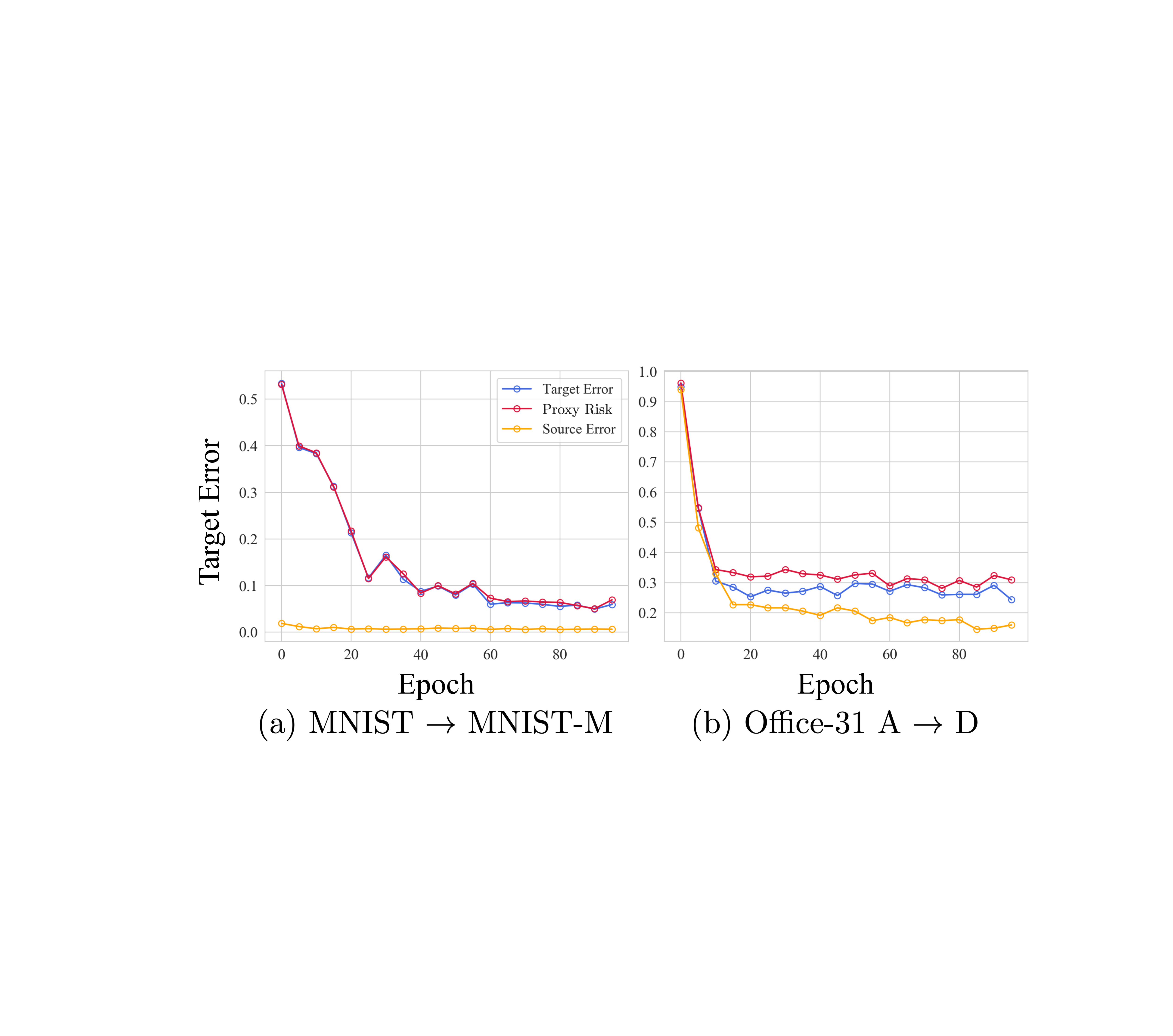}
\end{center}
\caption{Estimating stopping criteria: Target risk, source risk, and the proxy risk across training procedures.} \label{fig_stop}
\end{figure}

\paragraph{Estimating Stopping Criteria}
Without access to target labels, it is nontrivial to determine when to perform early stopping for DIR \citep{prechelt1998early}. Previous works leverage source error or self-labeled validation sets \citep{ganin2014unsupervised} to serve as stopping criterion. Figure \ref{fig_stop} shows the target error along with the source error and proxy error during the training of DANNs. The check model has the same architecture as the candidate model and the predictions are approximated with optimal division. In both experiments, proxy risks are well aligned with target errors. Notably, the proxy risk is almost the same as the target risk in MNIST$\rightarrow$MNIST-M. These results suggest that proxy risk is a good criterion for early stopping.

\subsection{Estimating Performance Drop of Supervised Learning under Domain Shift}\label{sec:est_supervised}
In this section, we aim to estimate the target risk of non-adaptive models that are trained only on the source, i.e., standard supervised learning. We compare our method with (1) \citet{ben2010theory}'s bound (Bound \eqreff{bound_ben}) and (2) a method based on confidence scores \citep{elsahar2019annotate}. For the first approach, we approximate the bound \eqreff{bound_ben} by estimating $R_{S}(h) + d_{\mathcal{H}\Delta\mathcal{H}}(p_{S}, p_{T})$ with $\mathcal{H} = \{ h \in \mathcal{H} | R_{S}(h) \leq \epsilon \}$, via an approach similar to that in Section \ref{sec_approx} (details are in the appendix). 
For the second approach, let $q_{h}(x)$ be the probability score (i.e., the max value of the softmax output) of hypothesis $h$ for $x \in \mathcal{X}$. \citet{elsahar2019annotate} compute confidence scores as the drop in average probability scores:
\begin{align}
    \textnormal{CONF}_{S,T}(h) = \mathbb{E}_{x \sim p_{S}}[q_{h}(x)] - \mathbb{E}_{x \sim p_{T}}[q_{h}(x)].
\end{align}
The target risk is then predicted by $R_{S}(h) + \textnormal{CONF}_{S,T}(h)$. We examine the methods on object classification, which contains 6 source/target pairs and digit classification, which has 12 source/target pairs after adding USPS. The architectures are fixed for the same task. To estimate risks on the new distribution, we consider domain-invariant classifiers with different divisions as check models. The encoder in the ``standard'' and ``complex'' check models have 4/6 and 6/12 layers, respectively, for digit/object classification. The check models and supervised prediction models share the same architecture.

\begin{figure*}[htbp]
\begin{center}   
\resizebox{2.1\columnwidth}{!}{
\includegraphics[width=2\linewidth]{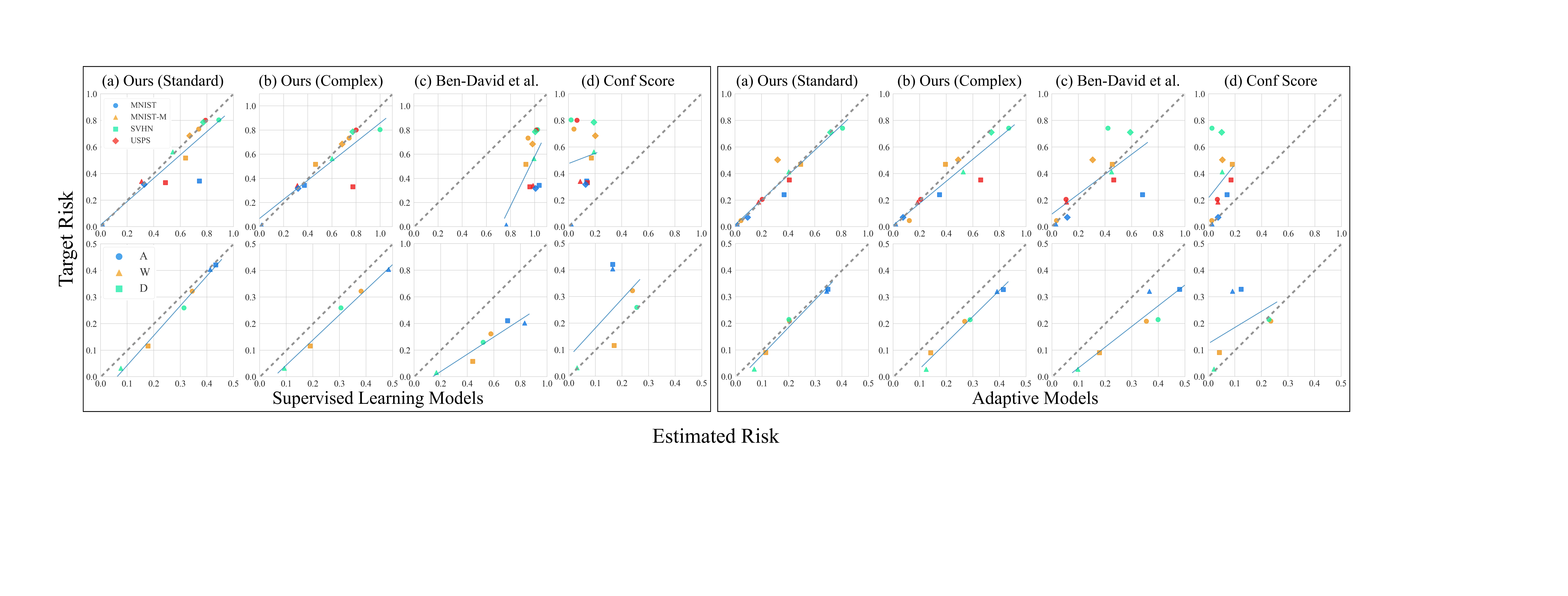}
}
\end{center}
\caption{Estimating performance drop. First row: Digit Classification, second row: object classification. The dashed line represents perfect prediction (target risk $=$ predicted risk). Shape and color of points indicates different source and target domain, respectively. Points beneath (above) the dashed line indicate overestimation (underestimation). The solid lines are regression lines. } 
\label{fig_combine}
\vspace{2mm}
\end{figure*}

The results in Figure \ref{fig_combine} (left) show that our methods consistently provide much better predictions than the baselines. With a complex encoder in the check model, we slightly overestimate the target risk, aligned with our theory: check models with properly controlled embedding complexity result in better prediction. \citet{ben2010theory}'s bound tends to overestimate the target risk, suggesting that the $\mathcal{H}\Delta\mathcal{H}$-divergence is too pessimistic for empirical estimation. In contrast, the confidence score approach largely underestimates the target risk. Table \ref{table_1} shows the quantitative results (SL-Digit and SL-Object): the average absolute error over domain pairs and the Pearson correlation coefficient between target risks and predictions. Our methods outperform baselines by a large margin in both metrics. 

\begin{table}[]
\begin{center}
\resizebox{\columnwidth}{!}{
\begin{tabular}{ |c|c|c|c|c|c|c| } 
 \hline
  Method& \multicolumn{2}{|c|}{Ours (Standard)} &  \multicolumn{2}{|c|}{Ben-David et al.} & \multicolumn{2}{|c|}{Conf Score}\\
 \hline\hline
 Metric & Err & PCC & Err & PCC & Err & PCC\\ 
 \hline\hline
 SL-Digit & \textbf{0.073} & \textbf{0.880} & 0.450& 0.585& 0.406& 0.103 \\ 
 SL-Object & \textbf{0.034} & \textbf{0.995} & 0.281& 0.945& 0.106& 0.551\\ 
  \hline\hline
 DIR-Digit & \textbf{0.043} & \textbf{0.957} & 0.124& 0.693& 0.242& 0.273 \\ 
 DIR-Object & \textbf{0.021} & \textbf{0.986} & 0.114& 0.932& 0.088& 0.459\\ 
 \hline
\end{tabular}
}
\end{center}
\caption{Estimating the target risk. We show the average error (lower is better) and Pearson correlation coefficient (higher is better) on different tasks. We estimate the target risk for supervised learning models (SL) and adaptive models (DIR).} \label{table_1}
\end{table}

\subsection{Estimating Adaptability between Domains}
In this section, we repeat the experiments in Section~\ref{sec:est_supervised}, this time for estimating the target risk of adaptive, domain-invariant classifiers (DANNs). Different from the previous section, we tighten \citet{ben2010theory}'s bound by setting $\mathcal{H} = \mathcal{P}_{\mathcal{FG}}^{\epsilon}$. Figure~\ref{fig_combine} (right) shows the results. Our method still consistently outperforms baselines in both tasks. Compared to estimating target risk for nonadaptive models, improvements in performance are observed for all methods, as Table~\ref{table_1} (DIR-Digit and DIR-Object) shows. Our methods produce estimates within a few percent of the true accuracy, while the baselines often underestimate (confidence score) or overestimate \citep{ben2010theory} the target risk.

\subsection{Error Detection}

Besides estimating the risk on new distributions, it is also important to know whether a prediction is reliable at a specific new test point. Our approach easily extends to predicting point-wise error, i.e., predicting misclassification at a target point $x^{\ast}$.
Recall that to approximate the proxy error of $h$, we train a check model $h^{\prime}$ to maximize the disagreement with $h$. We use $h^{\prime}$ to predict misclassification: if $h^{\prime}(x^{\ast}) \neq h(x^{\ast})$, then we should not trust $h(x^{\ast})$ and predict an error.

To evaluate this method quantitatively, we formulate error prediction as binary classification, and compute the F1 score of error detection on the target domain for supervised and adaptive models. The results, shown in Table \ref{table_2}, demonstrate that we can not only quantify the expected error in unseen distributions but also estimate the point-wise error accurately.

\begin{table}[H]
\vspace{-0mm}
\begin{center}
\resizebox{0.85\columnwidth}{!}{
\begin{tabular}{ |c|c|c|c|c| } 
 \hline
  &  \multicolumn{2}{|c|}{Digit Classification} & \multicolumn{2}{|c|}{Object Classification}\\
  \hline
  & M$\rightarrow$MM & S$\rightarrow$M & A$\rightarrow$D & A$\rightarrow$W \\
 \hline\hline
 SL & 0.985 & 0.908 & 0.925 & 0.938\\ 
 \hline
 DIR & 0.980 & 0.885 & 0.908 & 0.928\\ 
  \hline
\end{tabular}
}
\end{center}
\caption{F1 score of error detection on the target domain for supervised learning models (SL) and adaptive models (DIR).} \label{table_2}
\end{table}

\section{Conclusion}
In this paper, we made two contributions: (1) We leverage domain-invariant classifiers to empirically estimate the target risk, i.e., performance on a new, shifted, unlabeled dataset, of any given supervised or domain adaptation model. This approach applies to estimating risk on a data set for a single classifier, predicting point-wise error, and estimating the risk for a set of given classifiers, e.g., for model selection.
(2) To obtain good estimators, we theoretically and empirically analyze the effect of embedding complexity on the target risk in domain-invariant representations. We observe that the embedding complexity is an important factor for adaptability to the target distribution, much more than the complexity of the predictor part, and more than its roe for non-adaptive, supervised learning. 



Interesting directions of future work include adopting other domain adaptation algorithms as check models, and applying our approach to structured tasks, e.g., detection or segmentation.

\section*{Acknowledgements}

This work was supported by the MIT-IBM Watson AI Lab, NSF CAREER Award 1553284, an NSF BIGDATA award and the MIT-MSR TRAC collaboration. We thank Tongzhou Wang, Joshua Robinson, Wei Fang, Wei-Chiu Ma, and Chen-Ming Chuang for helpful discussions and suggestions.

\bibliography{example_paper}
\bibliographystyle{icml2020}

\appendix
\section{Proofs}

\subsection{Proof of Lemma 3}

\newtheorem{innercustomlemma}{Lemma}
\newenvironment{customlemma}[1]
  {\renewcommand\theinnercustomlemma{#1}\innercustomlemma}
  {\endinnercustomlemma}

\begin{customlemma}{3}\label{three}
Given a hypothesis class $\mathcal{P}$, for all $h \in \mathcal{H}$,
\begin{align}
R_{T}(h) 
&\leq \underbrace{\sup_{h^{\prime } \in \mathcal{P}}R_{T}(h, h^{\prime})}_{\textnormal{Proxy Risk}} + \underbrace{\inf_{h^{\prime} \in \mathcal{P}}R_{T}(h^{\prime})}_{\textnormal{Bias}}.
\label{bound5}
\end{align}
\end{customlemma}

\begin{proof}
Let $h^{\ast} = \arginf_{h \in \mathcal{P}}R_{T}(h)$, by the triangle inequality,
\begin{align}
    R_{T}(h) &\leq R_{T}(h, h^{\ast}) + R_{T}(h^{\ast}) \\
             &\leq \sup_{h^{\prime} \in \mathcal{P}}R_{T}(h, h^{\prime}) + \inf_{h^{\prime} \in \mathcal{P}}R_{T}(h^{\prime}).
\end{align}
\end{proof}

\subsection{Proof of Lemma 4}

\begin{customlemma}{4}\label{seven}
Given a hypothesis class $\mathcal{P}$, for all $h \in \mathcal{H}$,
\begin{align}
 \underbrace{|\sup_{h^{\prime} \in \mathcal{P}} R_{T}(h, h^{\prime}) - R_{T}(h)|}_{\textnormal{Estimation Error}} \leq \sup_{h^{\prime} \in \mathcal{P}}R_{T}(h^{\prime}).
\end{align}
\end{customlemma}

\begin{proof}
First, by Lemma \ref{three}, we have
\begin{align}
 \sup_{h^{\prime} \in \mathcal{P}_{\mathcal{FG}}^{\epsilon}}R_{T}(h, h^{\prime}) - R_{T}(h) \geq -\inf_{h^{\prime} \in \mathcal{P}_{\mathcal{FG}}^{\epsilon}}R_{T}(h^{\prime}).
 \label{upbound}
\end{align}

Next, by the triangle inequality,
\begin{align}
R_{T}(h, h^{\prime}) \leq R_{T}(h) + R_{T}(h^{\prime}).
\end{align}
The above inequality holds for all $h^{\prime} \in \mathcal{P}_{\mathcal{FG}}^{\epsilon}$, by taking supremum on both sides of the inequality, we have 
\begin{align}
\sup_{h^{\prime} \in \mathcal{P}_{\mathcal{FG}}^{\epsilon}} R_{T}(h, h^{\prime}) &\leq \sup_{h^{\prime} \in \mathcal{P}_{\mathcal{FG}}^{\epsilon}} (R_{T}(h) + R_{T}(h^{\prime})) \\
&= R_{T}(h) + \sup_{h^{\prime} \in \mathcal{P}_{\mathcal{FG}}^{\epsilon}}R_{T}(h^{\prime}),
\end{align}
implying that
\begin{align}
\sup_{h^{\prime} \in \mathcal{P}_{\mathcal{FG}}^{\epsilon}} R_{T}(h, h^{\prime}) - R_{T}(h) \leq \sup_{h^{\prime} \in \mathcal{P}_{\mathcal{FG}}^{\epsilon}}R_{T}(h^{\prime}).
 \label{lobound}
\end{align}

Since $\inf_{h^{\prime} \in \mathcal{P}_{\mathcal{FG}}^{\epsilon}}R_{T}(h^{\prime}) \leq \sup_{h^{\prime} \in \mathcal{P}_{\mathcal{FG}}^{\epsilon}}R_{T}(h^{\prime})$, combining \eqref{upbound} and \eqref{lobound} completes the proof.
\end{proof}

\subsection{Proof of Theorem 6}

\newtheorem{innercustomthm}{Theorem}
\newenvironment{customthm}[1]
  {\renewcommand\theinnercustomthm{#1}\innercustomthm}
  {\endinnercustomthm}

\begin{customthm}{6}\label{five}
 For all $f \in \mathcal F$ and $g \in \mathcal G$,
\begin{align}
R_{T}(fg) \leq R_{S}(fg) &+ \underbrace{d_{\mathcal{F}\Delta\mathcal{F}}(p_{S}^{g}(Z), p_{T}^{g}(Z))}_{\text{\textnormal{Latent Divergence}}}  \nonumber\\ &+ \underbrace{d_{\mathcal{F}_{\mathcal{G}\Delta\mathcal{G}}}(p_{S}, p_{T})}_{\substack{\textnormal{Embedding Complexity}}}+\lambda_{\mathcal F\mathcal{G}}(g). \label{bound4}
\end{align}
where $\lambda_{\mathcal F\mathcal{G}}(g)$ is a variant of the best in-class joint risk:
\begin{align}
    \lambda_{\mathcal F\mathcal{G}}(g) =  \inf_{f^{\prime} \in \mathcal{F}, g^{\prime} \in \mathcal{G}} 2R_{S}(f^{\prime}g) + R_{S}(f^{\prime}g^{\prime}) + R_{T}(f^{\prime}g^{\prime}). \nonumber
\end{align}
\end{customthm}
\begin{proof}

Define $f^{\ast}g^{\ast}$ as follows:
\begin{align}
    f^{\ast}g^{\ast} = \arginf_{f^{\prime} \in \mathcal{F}, g^{\prime} \in \mathcal{G}} 2R_{S}(f^{\prime}g) + R_{S}(f^{\prime}g^{\prime}) + R_{T}(f^{\prime}g^{\prime}).
\end{align}

By the triangle inequality,
\begin{align}
R_{T}(fg) &\leq R_{T}(f^{\ast}g^{\ast}) + R_{T}(fg, f^{\ast}g^{\ast}) \\
&\leq R_{T}(f^{\ast}g^{\ast}) + R_{T}(fg, f^{\ast}g) + R_{T}(f^{\ast}g, f^{\ast}g^{\ast}).
    \label{eq:1}
\end{align}
The second term in the R.H.S of Eq. \ref{eq:1} can be bounded as
\begin{align}
&\;\;\;\;\;R_{T}(fg, f^{\ast}g) \\&\leq  R_{S}(fg, f^{\ast}g) + |R_{S}(fg, f^{\ast}g) - R_{T}(fg, f^{\ast}g)| \\
&\leq R_{S}(fg, f^{\ast}g) + \sup_{f,f^{\prime} \in \mathcal{F}}|R_{S}(fg, f^{\prime}g) - R_{T}(fg, f^{\prime}g)| \\
&= R_{S}(fg, f^{\ast}g) + d_{\mathcal{F}\Delta\mathcal{F}}(p_{S}^{g}(Z), p_{T}^{g}(Z)) \\
&\leq R_{S}(fg) + R_{S}(f^{\ast}g) + d_{\mathcal{F}\Delta\mathcal{F}}(p_{S}^{g}(Z), p_{T}^{g}(Z)).
\end{align}
The last inequality follows from the triangle inequality. The third term in the R.H.S of Eq. \ref{eq:1} can be bounded similarly:
\begin{align}
&\;\;\;\;\;R_{T}(f^{\ast}g, f^{\ast}g^{\ast}) \\&\leq  R_{S}(f^{\ast}g, f^{\ast}g^{\ast}) + |R_{S}(f^{\ast}g, f^{\ast}g^{\ast}) - R_{T}(f^{\ast}g, f^{\ast}g^{\ast})| \\
&\leq R_{S}(f^{\ast}g, f^{\ast}g^{\ast}) \nonumber \\
&\;\;\;\;\;+\sup_{f \in \mathcal{F}, g, g^{\prime} \in \mathcal{G} }|R_{S}(f^{\prime}g, f^{\prime}g^{\prime}) - R_{T}(f^{\prime}g, f^{\prime}g^{\prime})| \\
&= R_{S}(f^{\ast}g, f^{\ast}g^{\ast}) + d_{\mathcal{F}_{\mathcal{G}\Delta\mathcal{G}}}(p_{S}(X), p_{T}(X)) \\
&\leq  R_{S}(f^{\ast}g) + R_{S}(f^{\ast}g^{\ast}) + d_{\mathcal{F}_{\mathcal{G}\Delta\mathcal{G}}}(p_{S}(X), p_{T}(X)).
\end{align}

Plugging the above bounds into \eqref{eq:1}, we have
\begin{align}
R_{T}(fg) \leq R_{S}(fg) &+ d_{\mathcal{F}\Delta\mathcal{F}}(p_{S}^{g}(Z), p_{T}^{g}(Z)) \\&+ d_{\mathcal{F}_{\mathcal{G}\Delta\mathcal{G}}}(p_{S}(X), p_{T}(X)) + \lambda_{\mathcal{FG}}(g),
\end{align}
where the $\lambda_{\mathcal{FG}}(g)$ emerges by
\begin{align}
\lambda_{\mathcal{FG}}(g) &= 2R_{S}(f^{\ast}g) + R_{S}(f^{\ast}g^{\ast}) + R_{T}(f^{\ast}g^{\ast}) \\
&= \inf_{f^{\prime} \in \mathcal{F}, g^{\prime} \in \mathcal{G}} 2R_{S}(f^{\prime}g) + R_{S}(f^{\prime}g^{\prime}) + R_{T}(f^{\prime}g^{\prime}).
\end{align}
\end{proof}

\newtheorem{innercustomproposition}{Proposition}
\newenvironment{customproposition}[1]
  {\renewcommand\theinnercustomproposition{#1}\innercustomproposition}
  {\endinnercustomproposition}

\subsection{Proof of Proposition 7}

\begin{customproposition}{7}\label{six}
In an $N$-layer feedforward neural network $h = f_{i}g_{i} \in \mathcal{F}_{i}\mathcal{G}_{i} = \mathcal{H}$, the following inequalities hold for all $i \leq j \leq N-1$:
\begin{align}
    d_{{\mathcal{F}_{i}}_{\mathcal{G}_{i}\Delta\mathcal{G}_{i}}}(p_{S}, p_{T}) &\leq
    d_{{\mathcal{F}_{j}}_{\mathcal{G}_{j}\Delta\mathcal{G}_{j}}}(p_{S}, p_{T})   \nonumber\\
         d_{{\mathcal{F}_{i}}\Delta\mathcal{F}_{i}}(p_{S}^{g_{i}}(Z), p_{T}^{g_{i}}(Z)) &\geq
    d_{{\mathcal{F}_{j}}\Delta\mathcal{F}_{j}}(p_{S}^{g_{j}}(Z), p_{T}^{g_{j}}(Z)).  \nonumber
\end{align}
\end{customproposition}

\begin{proof}
Recall that an $N$-layer feedforward neural network can be decomposed as $h = f_{i}g_{i} \in \mathcal{F}_{i}\mathcal{G}_{i} = \mathcal{H}$ for $i \in \{1, 2, \dots, N-1 \}$, where the embedding $g_{i}$ is formed by the first layer to the $i$-th layer and the predictor $f_{i}$ is formed by the $i+1$-th layer to the last layer. We define $q_{ij}: \mathcal{Z}_{i} \rightarrow \mathcal{Z}_{j}$ to be the function formed by the $i$-th layer to $j$-th layer, where $\mathcal{Z}_{k}$ is the latent space formed by the encoder $g_{k}: \mathcal{X} \rightarrow \mathcal{Z}_{k}$. We denote the class of $q_{ij}$ as $\mathcal{Q}_{ij}$.

We now prove the first inequality. By the definition of the $\mathcal{F}_{\mathcal{G}\Delta\mathcal{G}}$-divergence, for every $i \leq j$
\begin{align}
&d_{{\mathcal{F}_{i}}_{\mathcal{G}_{i}\Delta\mathcal{G}_{i}}}(p_{S}, p_{T}) \\
=& \sup_{\substack{f \in \mathcal{F}_{i} \\ g, g^{\prime} \in \mathcal{G}_{i}}} |R_{S}(fg, fg^{\prime}) - R_{T}(fg, fg^{\prime})| \\
=& \sup_{\substack{f \in \mathcal{F}_{j}, q \in \mathcal{Q}_{ij} \\ g, g^{\prime} \in \mathcal{G}_{i}}} |R_{S}(fqg, fqg^{\prime}) - R_{T}(fqg, fqg^{\prime})| \\
\leq& \sup_{\substack{f \in \mathcal{F}_{j} \\ q, q^{\prime} \in \mathcal{Q}_{ij} \\ g, g^{\prime} \in \mathcal{G}_{i}}} |R_{S}(fqg, fq^{\prime}g^{\prime}) - R_{T}(fqg, fq^{\prime}g^{\prime})| \\
=& \sup_{\substack{f \in \mathcal{F}_{j} \\ g, g^{\prime} \in \mathcal{G}_{j}}} |R_{S}(fg, fg^{\prime}) - R_{T}(fg, fg^{\prime})| \\
=&d_{{\mathcal{F}_{j}}_{\mathcal{G}_{j}\Delta\mathcal{G}_{j}}}(p_{S}, p_{T})
\end{align}

The second inequality can be proved similarly. By the definition of the $\mathcal{F}\Delta\mathcal{F}$-divergence, for every $i \leq j$
\begin{align}
&d_{\mathcal{F}_{j}\Delta\mathcal{F}_{j}}(p_{S}^{g_{j}}(Z), p_{T}^{g_{j}}(Z)) \\=& \sup_{f, f^{\prime} \in \mathcal{F}_{j}} |R_{S}(fg_{j}, f^{\prime}g_{j}) - R_{T}(fg_{j}, f^{\prime}g_{j})| \\
=& \sup_{f, f^{\prime} \in \mathcal{F}_{j}} |R_{S}(fq_{ij}g_{i}, f^{\prime}q_{ij}g_{i}) - R_{T}(fq_{ij}g_{i}, f^{\prime}q_{ij}g_{i})| \\
\leq&  \sup_{\substack{q \in \mathcal{Q}_{ij} \\ f, f^{\prime} \in \mathcal{F}_{j}}} |R_{S}(fqg_{i}, f^{\prime}qg_{i}) - R_{T}(fqg_{i}, f^{\prime}qg_{i})| \\
\leq&  \sup_{\substack{q, q^{\prime} \in \mathcal{Q}_{ij} \\ f, f^{\prime} \in \mathcal{F}_{j}}} |R_{S}(fqg_{i}, f^{\prime}q^{\prime}g_{i}) - R_{T}(fqg_{i}, f^{\prime}q^{\prime}g_{i})| \\
=& \sup_{f, f^{\prime} \in \mathcal{F}_{i}} |R_{S}(fg_{i}, f^{\prime}g_{i}) - R_{T}(fg_{i}, f^{\prime}g_{i})| \\
=& d_{\mathcal{F}_{i}\Delta\mathcal{F}_{i}}(p_{S}^{g_{i}}(Z), p_{T}^{g_{i}}(Z))
\end{align}
\end{proof}

\section{Experiments}

\subsection{The Role of Inductive Bias}
\begin{wrapfigure}{r}{0.23\textwidth}
  \begin{center}
    \includegraphics[width=0.23\textwidth]{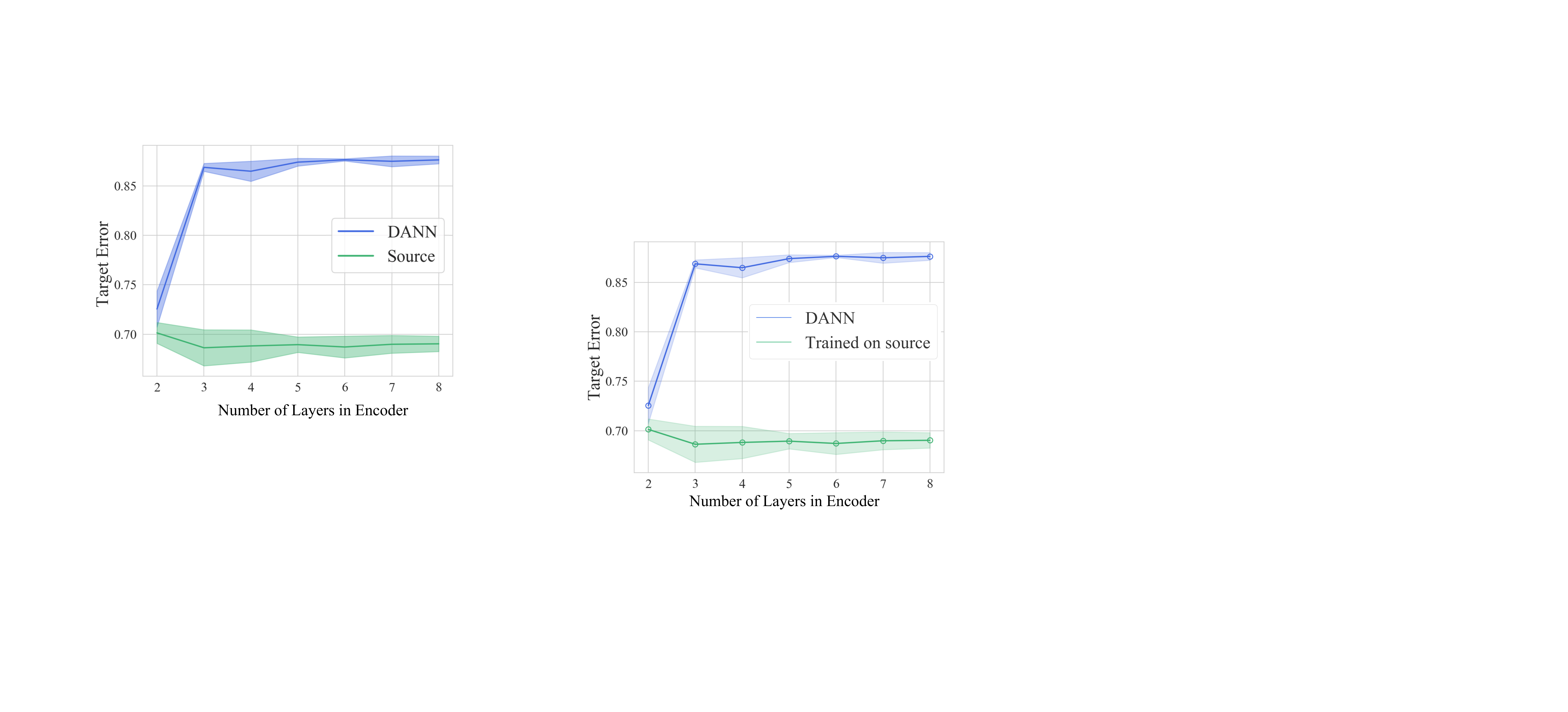}
  \end{center}
      \vspace{-4mm}
  \caption{The role of inductive bias: DANN with fully connected layers instead of a CNN.} \label{fig_ib}
      \vspace{-2mm}
\end{wrapfigure}

\begin{figure}[t]
    \includegraphics[width=\linewidth]{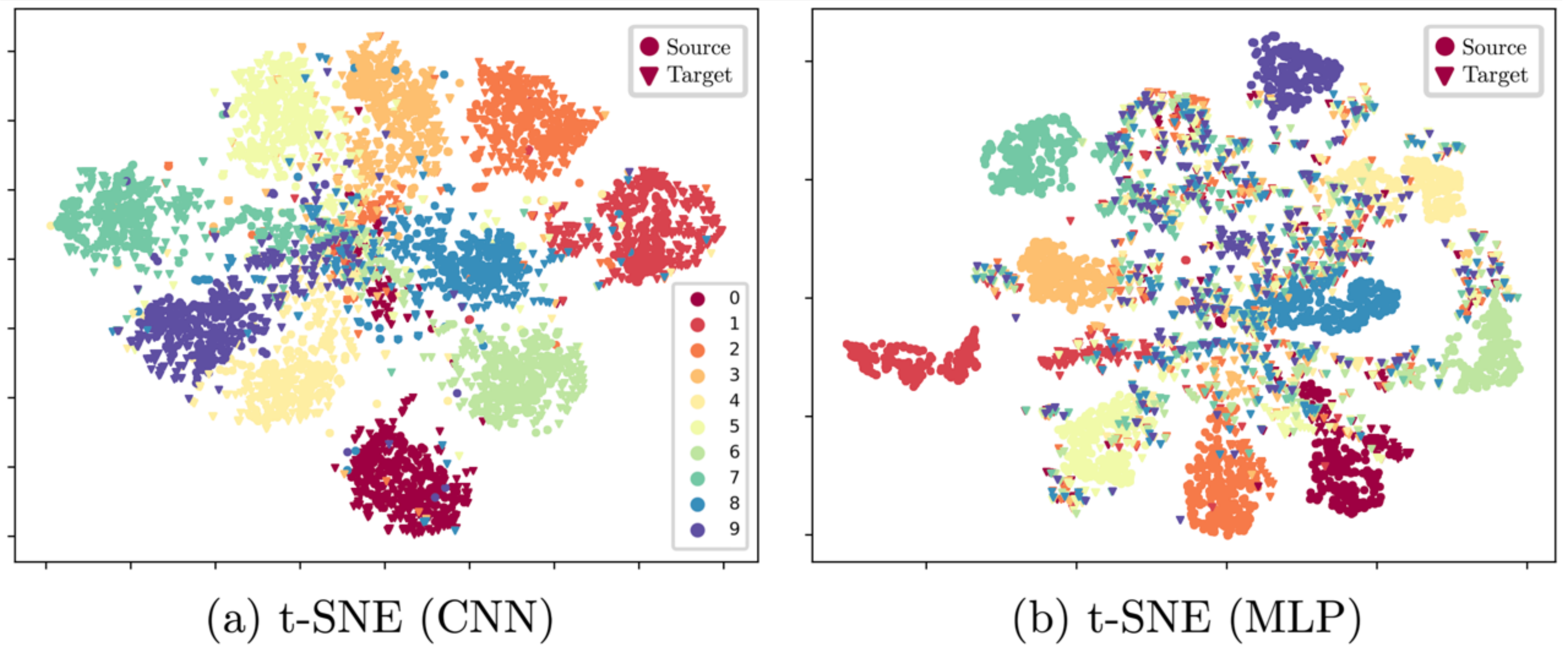}
    \vspace{-4mm}
    \caption{
       t-SNE projections of representations with different inductive biases. CNN encoders result in source and target representations that are well aligned. In contrast, MLP encoders lose label-consistency while minimizing the latent divergence between domains.
    } \label{fig_tsne}
  \vspace{-0mm}
\end{figure}

Besides the number of layers and number of hidden neurons, we investigate the importance of inductive bias for domain invariant representations, by replacing the CNN encoder with an MLP. The results for M$\rightarrow$M-M are shown in Figure \ref{fig_ib}. The target error with the MLP encoder is significantly higher than with a CNN encoder. 
Even more, with respect to target error, the model is very sensitive to its complexity. Increasing the number of layers worsens the performance, and is worse than not doing any domain adaptation. 
To gain deeper insight,  in Figure \ref{fig_tsne}, we use a t-SNE (\citet{maaten2008visualizing}) projection to visualize source and target distributions in the latent space. With the inductive bias of CNNs, the representations of target domain well align with those in the source domain. However, despite the overlap between source and target domains, the MLP encoder results in serious label-mismatch.

\subsection{Effect of Embedding Complexity}

\begin{figure}[t]
    \includegraphics[width=\linewidth]{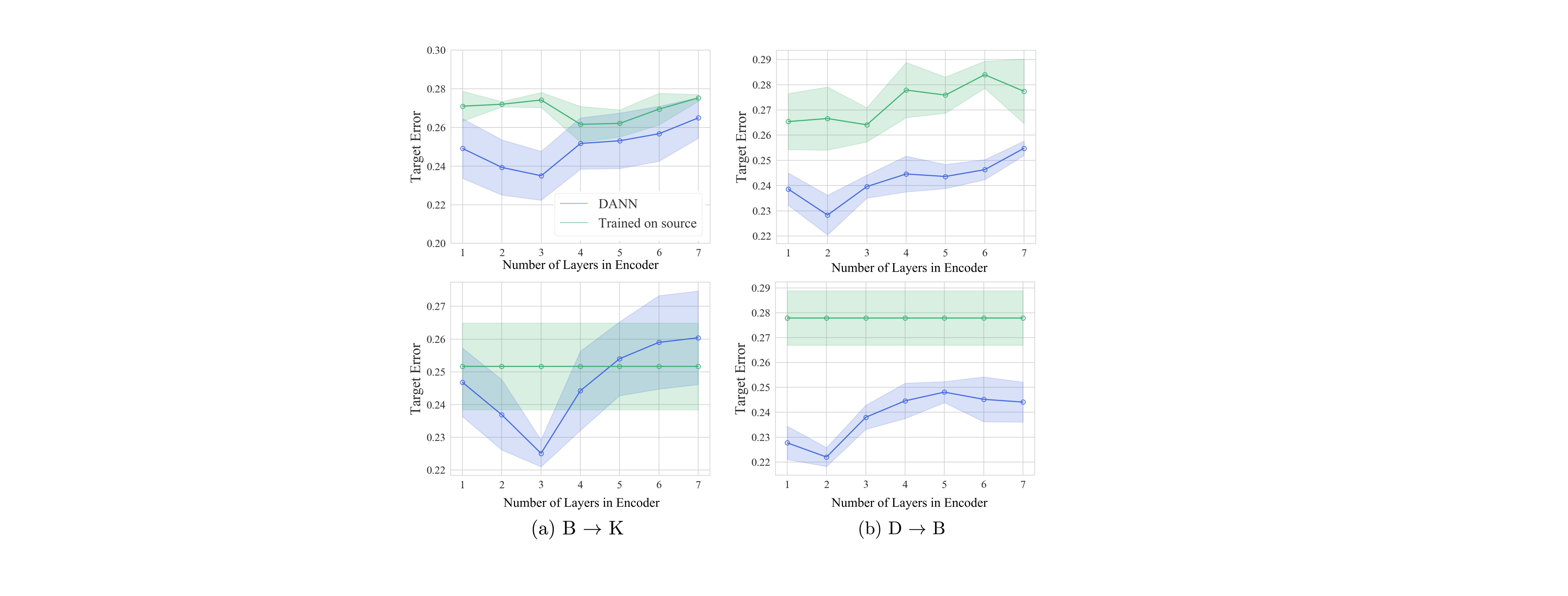}
    \vspace{-4mm}
    \caption{
       Amazon reviews dataset. First row: Fixed predictor class, varying number of layers in the encoder. Second row: Fixed total number of layers and optimizing domain-invariant loss in a single intermediate layer.
    } \label{fig_appendix_amazon}
  \vspace{-3mm}
\end{figure}

\paragraph{Sentiment Classification}
In addition to K$\rightarrow$B and D$\rightarrow$B, we show the results for B$\rightarrow$K and D$\rightarrow$B in Figure \ref{fig_appendix_amazon}. In agreement with the results in the main paper, the target error decreases initially, and then increases as more layers are added to the encoder. 

\begin{figure}[t]
    \includegraphics[width=\linewidth]{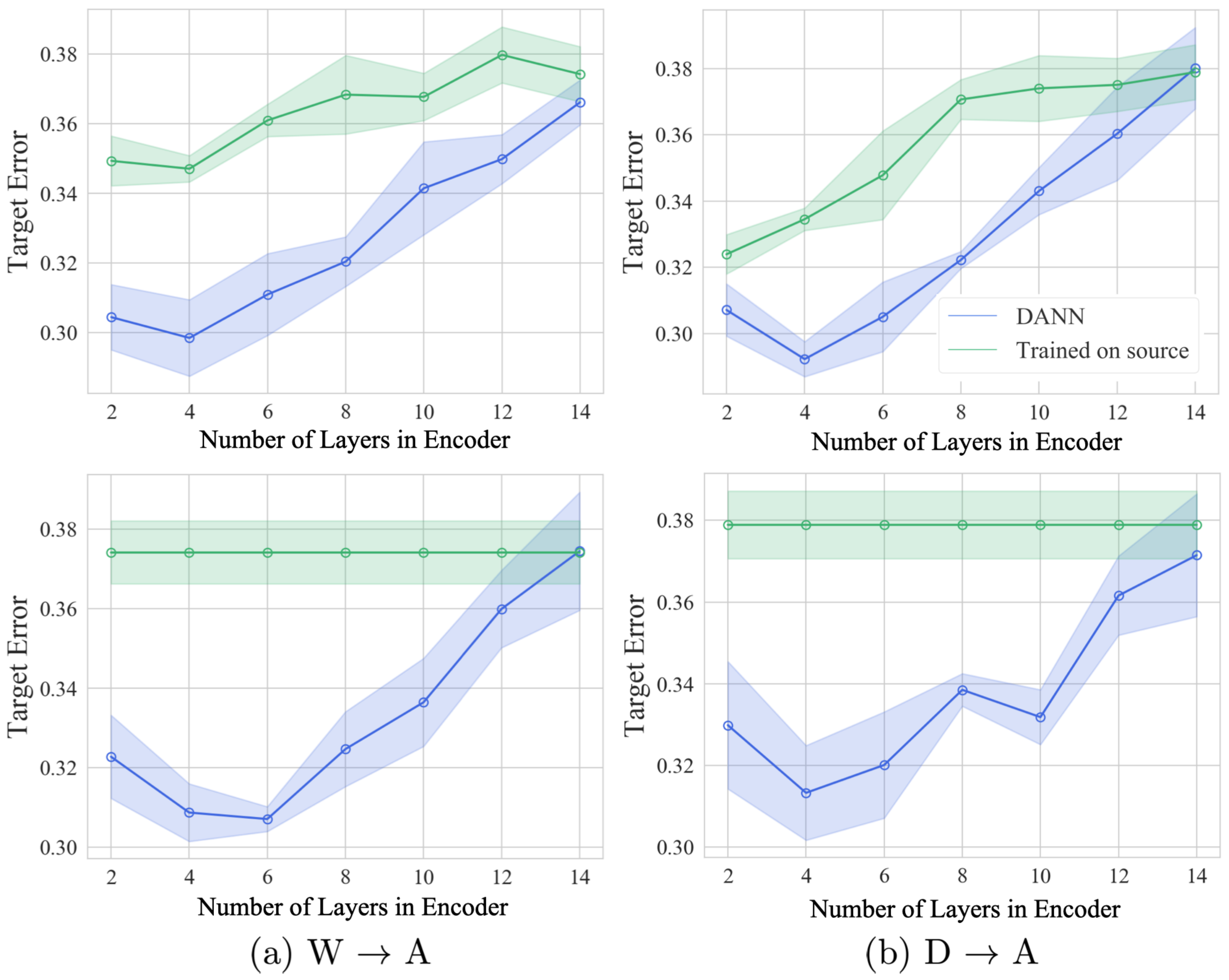}
    \vspace{-4mm}
    \caption{
       Office-31 dataset. First row: Fixed predictor class, varying number of layers in the encoder. Second row: Fixed total number of layers and optimizing domain-invariant loss in a single intermediate layer.
    } \label{fig_appendix_office}
  \vspace{-3mm}
\end{figure}

\paragraph{Object Classification}
In addition to A$\rightarrow$W and A$\rightarrow$D, we show the results for W$\rightarrow$A and D$\rightarrow$A in Figure \ref{fig_appendix_office}. Again, the target error decreases initially and increase as the encoder becomes more complex.

\subsection{Effect of Predictor Complexity}
\label{a_pred_trad}

\begin{figure}[t]
    \includegraphics[width=\linewidth]{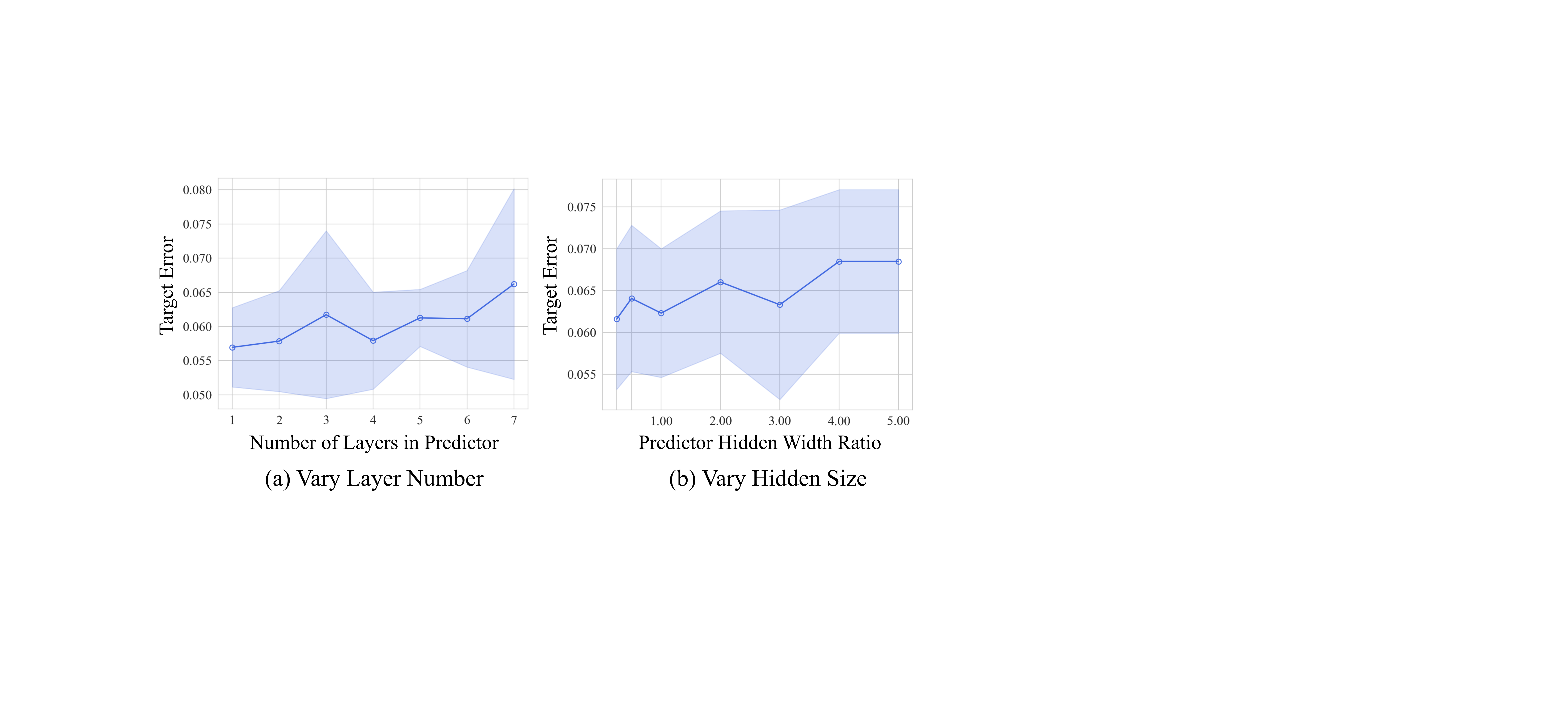}
    \vspace{-5mm}
    \caption{
       Effect of predictor complexity on MNIST$\rightarrow$MNIST-M.  (a) Fix the encoder class and vary the number of layers in the predictor. (b) Fix the encoder class and vary the hidden width of the predictor.
    } \label{fig_pt}
  \vspace{-4mm}
\end{figure}

Next, we investigate the effect of predictor complexity with the MNIST$\rightarrow$MNIST-M data. Following the procedure in Section 5, we augment the original predictor with 1 to 7 additional CNN layers and fix the number of layers in the encoder to $4$ or vary the hidden width of the predictor. The results are shown in Figure \ref{fig_pt}. The target error slightly decreases as the number of layers in the predictor increases, but much less than the $19.8\%$ performance drop when we vary the number of layers in the \emph{encoder} (Section 5.2, see also Fig.~\ref{fig_appendix_amazon} and \ref{fig_appendix_office}): when augmenting the predictor with 7 layers, the target error only decreases by $0.9\%$. Therefore, we focus on the embedding complexity in the main paper.

\section{Computing \citet{ben2010theory}'s Bound}
In Section 7.3 and 7.4, we estimate \citet{ben2010theory}'s bound with different hypothesis class $\mathcal{H}$. To justify the choice of $\mathcal{H}$, we have to trace back to the proof of Theorem~2 \citep{ben2010theory}.

\begin{customthm}{2}\label{two}  \textnormal{\citep{ben2010theory}} 
  For all hypotheses $h \in \mathcal H$, the target risk is bounded as
\begin{align}
R_{T}(h) \leq R_{S}(h) + d_{\mathcal{H}\Delta \mathcal{H}}(p_{S}, p_{T}) + \lambda_{\mathcal H},
\label{bound_ben}
\end{align}
where $\lambda_{\mathcal H}$ is the best joint risk 
\begin{align*}
\lambda_{\mathcal H} \coloneqq \inf_{h' \in \mathcal{H}}[R_S(h') + R_T(h')].
\end{align*}
\end{customthm}
\begin{proof}
Define the optimal hypothesis $h^{\ast}$:
\begin{align}
    h^{\ast} = \arginf_{h^{\prime} \in \mathcal{H}} R_{S}(h^{\prime}) + R_{T}(h^{\prime})
\end{align}
We then bound the target risk of $h$ as follows:
\begin{align}
&\;\;\;\;\;R_{T}(h) \\&\leq R_{T}(h, h^{\ast}) + R_{T}(h^{\ast}) \\
&\leq |R_{S}(h, h^{\ast}) - R_{T}(h, h^{\ast})| + R_{S}(h, h^{\ast}) + R_{T}(h^{\ast})  \\
&\leq R_{S}(h) + |R_{S}(h, h^{\ast}) - R_{T}(h, h^{\ast})| \nonumber \\&\qquad\qquad\qquad\qquad\qquad+ R_{S}(h^{\ast}) + R_{T}(h^{\ast})\\
&= R_{S}(h) + |R_{S}(h, h^{\ast}) - R_{T}(h, h^{\ast})| + \lambda_{\mathcal H}\\
&\leq R_{S}(h) + \underbrace{\sup_{h, h^{\prime} \in \mathcal{H}}|R_{S}(h, h^{\prime}) - R_{T}(h, h^{\prime})|}_{\textnormal{$\mathcal{H}\Delta\mathcal{H}$-Divergence}} + \lambda_{\mathcal H} \label{hdh_ineq}
\end{align}
where the first and third inequality follow from the triangle inequality and we replace $h^{\ast}$ with $h^{\prime}$ in the last inequality.
\end{proof}

The tightness of the bound depends on the hypothesis class $\mathcal{H}$ that the $\mathcal{H}\Delta\mathcal{H}$-divergence uses. According to \eqref{hdh_ineq} in the proof, to take the supremum, we have to ensure that the candidate hypothesis $h$ and the optimal hypothesis $h^{\ast}$ belong to $\mathcal{H}$.

For instance, a model $h$ pretrained on the source belongs to $\mathcal{H} = \{ h \in \mathcal{H} | \hat{R}_{S}(h) \leq \epsilon \}$ but not necessarily $\mathcal{P}_{\mathcal{FG}}^{\epsilon}$. 
However, to estimate the target risk for a domain-invariant classifier $h = fg$ which satisfies $R_{S}(h) + \alpha d(p_{S}^{g}(Z), p_{T}^{g}(Z) \leq \epsilon$ (Section 7.4), we can tighten the bound by setting $\mathcal{H} = \mathcal{P}_{\mathcal{FG}}^{\epsilon}$ since $h \in \mathcal{P}_{\mathcal{FG}}^{\epsilon}$. Note that in return, the unobserved best joint risk increases from $\lambda_{\mathcal H}$ to $\lambda_{\mathcal{P}_{\mathcal{FG}}^{\epsilon}}$.

We adopt the computational approach described in Section 7.2 to approximate the $\mathcal{H}\Delta\mathcal{H}$-divergence. For instance, to approximate $\mathcal{P}_{\mathcal{FG}}^{\epsilon}\Delta\mathcal{P}_{\mathcal{FG}}^{\epsilon}$-divergence, we optimize the following objective:
\begin{align}
    \max_{fg, f^{\prime}g^{\prime} \in \mathcal{FG}} &R_{T}(fg, f^{\prime}g^{\prime}) - R_{S}(fg, f^{\prime}g^{\prime}) \nonumber \\ &- \lambda( R_{S}(fg) + R_{S}(f^{\prime}g^{\prime})) \nonumber \\
    &- \lambda\alpha ( d(p_{S}^{g}(Z), p_{T}^{g}(Z)) + d(p_{S}^{g^{\prime}}(Z), p_{T}^{g^{\prime}}(Z)))
\end{align}
where $\lambda > 0$. We empirically estimate the $R_T$ and $R_S$ with the validation set and minimize the objective via standard stochastic gradient descent.

\section{Experiment Details and Network Architectures}
\label{a_arch}

\subsection{Amazon Review Dataset}
\label{a_ard}
The learning rate of the Adam optimizer is set to $1 \times e^{-3}$ and the model is trained for 50 epochs. We adopt the original progressive training strategy for the discriminator \citep{ganin2016domain} where the weight $\alpha$ for the domain-invariant loss is initiated at $0$ and is gradually changed to $1$ using the following schedule:
\begin{align}
    \alpha = \frac{2}{1 + \exp(-10 \cdot p)} - 1,
\end{align}
where $p$ is the training progress linearly changing from $0$ to $1$. The architecture of the hypothesis and discriminator are as follows:

\begin{center}
 \begin{tabular}{||c||} 
 \hline
 Encoder  \\ [0.5ex] 
 \hline\hline
 nn.Linear(5000, 128)  \\ 
 \hline
 nn.ReLU  \\
 \hline \hline
 nn.Linear(128, 128)  \\
 \hline
 nn.ReLU \\
 \hline
 $\times n$ (depends on the number of layers) \\
 \hline
\end{tabular}
\quad
 \begin{tabular}{||c||} 
 \hline
 Predictor  \\ [0.5ex] 
 \hline\hline
 nn.Linear(128, 128)  \\ 
 \hline
 nn.ReLU  \\
 \hline
 $\times n$ (depends on the number of layers) \\
 \hline \hline
 nn.Linear(128, 2)  \\
 \hline
 nn.Softmax \\
 \hline
\end{tabular}
\end{center}

\begin{center}
 \begin{tabular}{||c||} 
 \hline
 Discriminator  \\ [0.5ex] 
 \hline\hline
 nn.Linear(128, 256)  \\ 
 \hline
 nn.ReLU  \\
 \hline\hline
  nn.Linear(256, 256)  \\ 
 \hline
 nn.ReLU  \\
 \hline
 $\times$5\\
 \hline\hline
 nn.Linear(256, 2)  \\
 \hline
 nn.Softmax \\
 \hline
\end{tabular}
\end{center}

\subsection{Digit Classification}
The learning rate of the Adam optimizer is set to $1 \times e^{-3}$ and the model is trained for 100 epochs. The weight $\alpha$ for domain-invariant loss is initiated at $0$ and is gradually changed to $0.1$ using the same schedule in Section \ref{a_ard}. The tradeoff parameter $\lambda$ for computing proxy error is set to $50$ for all tasks. The architecture of the hypothesis and discriminator are as follows:

\begin{center}
 \begin{tabular}{||c||} 
 \hline
 Encoder  \\ [0.5ex] 
 \hline\hline
 nn.Conv2d(3, 64, kernel$\_$size=5)  \\ 
 \hline
 nn.BatchNorm2d \\
 \hline
 nn.MaxPool2d(2)  \\
 \hline 
 nn.ReLU \\
 \hline
 nn.Conv2d(64, 128, kernel$\_$size=5)  \\ 
 \hline
 nn.BatchNorm2d \\
 \hline
 nn.Dropout2d (only added for MNIST$\rightarrow$MNIST-M)\\
 \hline
 nn.MaxPool2d(2)  \\
 \hline 
 nn.ReLU \\
 \hline \hline
 nn.Conv2d(128, 128, kernel$\_$size=3, padding=1)  \\
 \hline
 nn.BatchNorm2d \\
 \hline
 nn.ReLU \\
 \hline
 $\times n$ (depends on the number of layers) \\
 \hline
\end{tabular}
\end{center}

\begin{center}
 \begin{tabular}{||c||} 
 \hline
 Predictor  \\ [0.5ex] 
 \hline\hline
 nn.Conv2d(128, 128, kernel$\_$size=3, padding=1)  \\
 \hline
 nn.BatchNorm2d \\
 \hline
 nn.ReLU \\
 \hline
 $\times n$ (depends on the number of layers) \\
 \hline\hline
 flatten \\
 \hline
 nn.Linear(2048, 256)  \\ 
 \hline 
 nn.BatchNorm1d \\
 \hline
 nn.ReLU  \\
 \hline
 nn.Linear(256, 10) \\
 \hline 
 nn.Softmax \\
 \hline
\end{tabular}
\end{center}

\begin{center}
 \begin{tabular}{||c||} 
 \hline
 Discriminator  \\ [0.5ex] 
 \hline\hline
 nn.Conv2d(128, 256, kernel$\_$size=3, padding=1)  \\
 \hline
 nn.ReLU \\
 \hline
 \hline\hline
 nn.Conv2d(256, 256, kernel$\_$size=3, padding=1)  \\
 \hline
 nn.ReLU \\
 \hline
 $\times 4$ \\
 \hline\hline
 Flatten \\
 \hline
 nn.Linear(4096, 512)  \\ 
 \hline 
 nn.ReLU  \\
 \hline
 nn.Linear(512, 512)  \\ 
 \hline 
 nn.ReLU  \\
 \hline 
 nn.Linear(512, 2) \\
 \hline 
 nn.Softmax \\
 \hline
\end{tabular}
\end{center}

In the hidden width experiments, we use the architectures above as a basis and scale their relevant their hidden widths.

\subsection{Office-31}
We exploit the ResNet-50 \citep{he2016deep} trained on ImageNet \citep{deng2009imagenet} for feature extraction.
The learning rate of the Adam optimizer is set to $3 \times e^{-4}$ and the models are trained for 100 epochs. The weight $\alpha$ for the domain-invariant loss is initiated at $0$ and is gradually changed to $1$ using the same schedule in Section \ref{a_ard}. The tradeoff parameter $\lambda$ for computing proxy error is set to $50$ for all tasks. The architecture of the hypothesis and discriminator are as follows:

\begin{center}
 \begin{tabular}{||c||} 
 \hline
 Encoder  \\ [0.5ex] 
 \hline\hline
 nn.Linear(2048, 256)  \\ 
 \hline
 nn.ReLU  \\
 \hline \hline
 nn.Linear(256, 256)  \\
 \hline
 nn.ReLU \\
 \hline
 $\times n$ (depends on the number of layers) \\
 \hline
\end{tabular}
\quad
 \begin{tabular}{||c||} 
 \hline
 Predictor  \\ [0.5ex] 
 \hline\hline
 nn.Linear(256, 256)  \\ 
 \hline
 nn.ReLU  \\
 \hline
 $\times n$ (depends on the number of layers) \\
 \hline \hline
 nn.Linear(256, 2)  \\
 \hline
 nn.Softmax \\
 \hline
\end{tabular}
\end{center}

\begin{center}
 \begin{tabular}{||c||} 
 \hline
 Discriminator  \\ [0.5ex] 
 \hline\hline
  nn.Linear(256, 256)  \\ 
 \hline
 nn.ReLU  \\
 \hline
 $\times$6\\
 \hline\hline
 nn.Linear(256, 2)  \\
 \hline
 nn.Softmax \\
 \hline
\end{tabular}
\end{center}

\end{document}